\documentclass{article}

\PassOptionsToPackage{numbers, compress}{natbib}

\usepackage[preprint]{neurips_2025}

\usepackage[utf8]{inputenc} 
\usepackage[T1]{fontenc}    
\usepackage{hyperref}       
\usepackage{url}            
\usepackage{booktabs}       
\usepackage{amsfonts}       
\usepackage{nicefrac}       
\usepackage{microtype}      
\usepackage{xcolor}         
\usepackage{graphics}
\usepackage{graphicx}
\usepackage{epsfig} 
\usepackage{mathptmx} 
\usepackage{times} 
\usepackage{amsmath} 
\usepackage{amssymb}  
\usepackage{float}
\usepackage[linesnumbered,ruled,vlined,commentsnumbered]{algorithm2e}
\usepackage{tabularx}
\usepackage{enumitem}
\usepackage{caption}
\usepackage{subcaption}
\usepackage{makecell}

\usepackage{chatbubbles}

\title{SD-OVON: A Semantics-aware Dataset and Benchmark Generation Pipeline for Open-Vocabulary Object Navigation in Dynamic Scenes}

\author{
    Dicong Qiu$^{1*}$, Jiadi You$^{1}$\thanks{Equal contribution.} , Zeying Gong$^{1}$, Ronghe Qiu$^{1}$, Hui Xiong$^{1,2\dagger}$ and Junwei Liang$^{1,2}$\thanks{Corresponding authors.}\\
    $^{1}$The Hong Kong University of Science and Technology (Guangzhou)\\
    $^{2}$The Hong Kong University of Science and Technology\\
    {\tt\small dqiu570@connect.hkust-gz.edu.cn}
}

\begin{document}

\maketitle

\begin{abstract}

We present the Semantics-aware Dataset and Benchmark Generation Pipeline for Open-vocabulary Object Navigation in Dynamic Scenes (SD-OVON). It utilizes pretraining multimodal foundation models to generate infinite unique photo-realistic scene variants that adhere to real-world semantics and daily commonsense for the training and the evaluation of navigation agents, accompanied with a plugin for generating object navigation task episodes compatible to the Habitat simulator. 
In addition, we offer two pre-generated object navigation task datasets, SD-OVON-3k and SD-OVON-10k, comprising respectively about 3k and 10k episodes of the open-vocabulary object navigation task, derived from the SD-OVON-Scenes dataset with 2.5k photo-realistic scans of real-world environments and the SD-OVON-Objects dataset with 0.9k manually inspected scanned and artist-created manipulatable object models.
Unlike prior datasets limited to static environments, SD-OVON covers dynamic scenes and manipulatable objects, facilitating both real-to-sim and sim-to-real robotic applications. This approach enhances the realism of navigation tasks, the training and the evaluation of open-vocabulary object navigation agents in complex settings. 
To demonstrate the effectiveness of our pipeline and datasets, we propose two baselines and evaluate them along with state-of-the-art baselines on SD-OVON-3k.
The datasets, benchmark and source code are publicly available at
\tt\href{https://sd-ovon.github.io}{https://sd-ovon.github.io}
\end{abstract}

\section{Introduction}
\label{sec:Introduction}

\begin{figure*}[ht]
    \centering
    \includegraphics[width=1.0\linewidth]{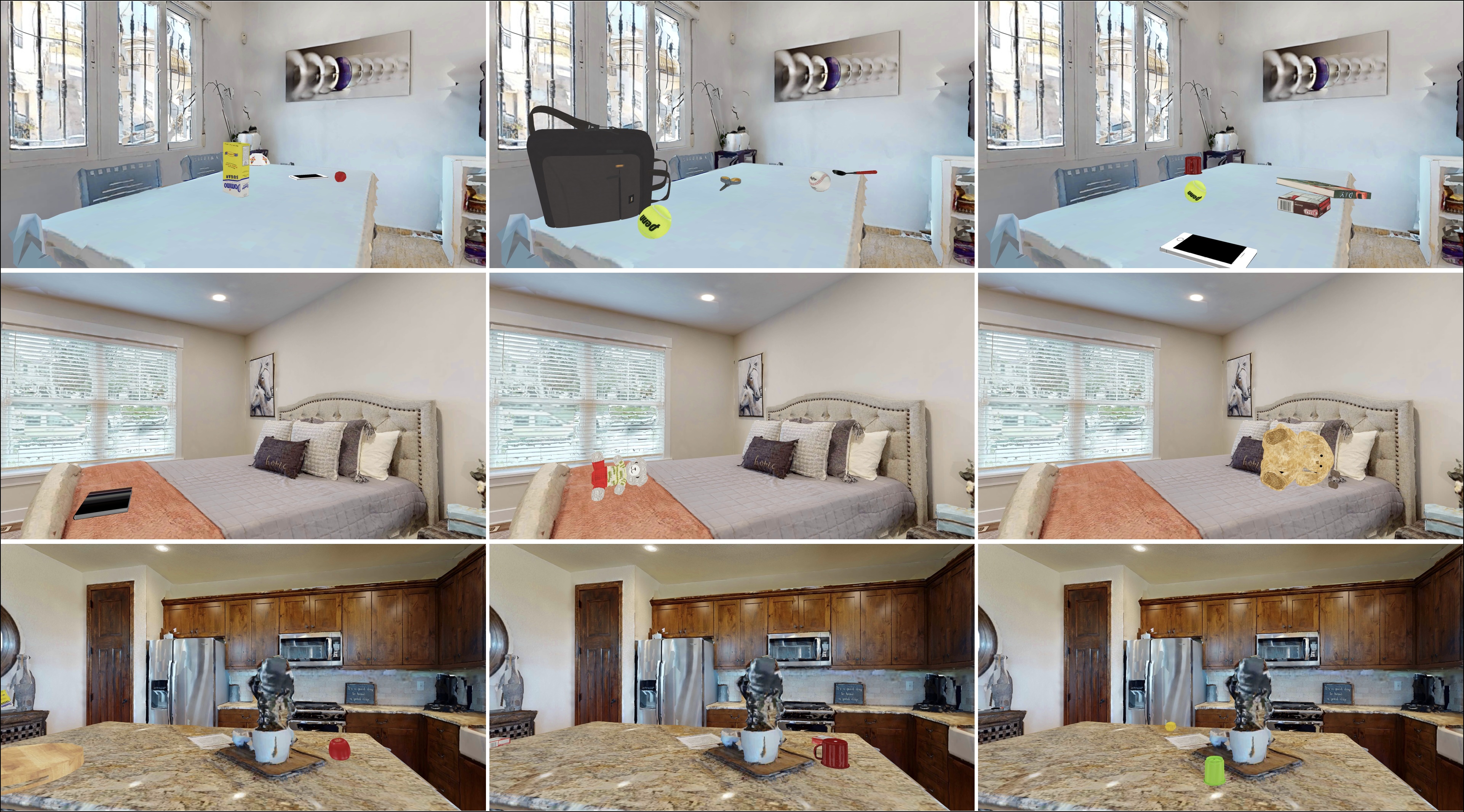}
    \caption{Visualization of example scene variants generated by SD-OVON. Manipulatable objects are placed in accordance to daily commonsense, considering both receptacle types and regions.}
    \label{fig:scene_variants}
\end{figure*}

Open-vocabulary object navigation in dynamic environments is a crucial capability that enables robots to function effectively in complex real-world environments. It presents a unique challenge in object goal navigation (ObjectNav)~\cite{anderson2018evaluation} necessitating systems capable of understanding and interacting with both unseen objects and complex environments that change over time. In the contemporary landscape of ObjectNav datasets and benchmarks, static environments prevail, presupposing that the composition of a scene remains constant and that objects are immovable fixtures within these settings~\cite{yokoyama2024ovon, ramakrishnan2021hm3d, dai2017scannet}. 
This limitation contradicts the dynamic nature of real-world environments, where objects used in daily basis are subject to frequent rearrangement, although furniture usually remains where it is.
Moreover, traditional ObjectNav task assumes a navigation agent possesses no prior knowledge of the environment, which not only diminishes the reusability of an environment, but also fails to reflect our daily situations, where humans frequently navigate familiar environments, such as home and office, that dynamically evolve--new items may appear, some may be removed, and others may be relocated, in accordance to people's daily habits and societal norms.

To address this gap, we propose \textit{SD-OVON}, a procedural approach that leverages state-of-the-art pretraining visual-language models (VLMs) and large language models (LLMs) to generate photo-realistic dynamic scenes from scans of real-world environments and manipulatable objects, adhering to real-world semantics and daily commonsense knowledge. Our approach enables the integration of real-world environment scans with both scans of real-world objects and artist-created 3D object models that are movable and manipulatable. Such an integration allows for the generation of a diverse array of scene variants derived from the aforementioned environment and objects, significantly enhancing the number of scenes available for training and evaluation of object navigation agents, particularly those with open-vocabulary capabilities. In contrast to traditional procedural scene generation methods that relies purely on artist-created 3D models~\cite{deitke2022ProcTHOR}, our methodology incorporates scans of real-world environments and objects, facilitating real-to-sim transferability and enhancing the realism of our dataset and benchmark.

With the capability of generating infinite unique photo-realistic scene variants and ObjectNav task episodes, SD-OVON significantly expands the landscape of available data for training and evaluating navigation agents. Additionally, we integrate the SD-OVON with Habitat simulator \cite{szot2021habitat} and offer 13k pre-generated and manually inspected ObjectNav task episodes derived from both photo-realistic scans and synthetic models of environments and movable objects, in the accompanied datasets, \textit{SD-OVON-3k} and \textit{SD-OVON-10k}. Two baselines, the \textit{Random Receptacle Navigation A*} and the \textit{Semantic Navigation A*}, are proposed and evaluated along with state-of-the-art open-vocabulary object navigation baselines on SD-OVON-3k, demonstrating the effectiveness of our proposed pipeline and the accompanied datasets on the open-vocabulary object navigation task. This work not only enhances the realism of navigation tasks but also allows for a more comprehensive and diverse evaluation of open-vocabulary navigation algorithms in dynamic and variable environments, thereby setting the stage for breakthroughs in open-vocabulary object navigation.

\section{Related Work}
\label{sec:RelatedWork}

\subsection{Open-Vocabulary Object Navigation}

Open-vocabulary object navigation (OVON) requires a robot to explore diverse, even unseen objects and environments and locate target objects based on natural language instructions, without being limited to a predefined set of object categories~\cite{wu2024towards}, as seen in the traditional object goal navigation \cite{anderson2018evaluation} task. Early works in vision-and-language navigation (VLN) laid the groundwork for robotic agents to follow natural language instructions in static environments \cite{anderson2018vision}. Recent advancements in LLMs~\cite{touvron2023llama, achiam2023gpt} and VLMs \cite{ren2024grounded, chen2024internvl, bai2025qwen2} have enabled robots to navigate dynamic environments with impressive zero-shot capabilities. Current research is divided into two main approaches: modular and end-to-end learning frameworks. Modular approaches
combine geometric mapping with VLM/LLM-based semantic exploration~\cite{yokoyama2024vlfm, yu2023l3mvn, qiu2024open},
align text prompts with visual embeddings for open-world navigation~\cite{ye2021auxiliary}, 
guiding an agent towards language-specified targets~\cite{zhang20233d, ramakrishnan2022poni, chaplot2020object}.
On the other hand, end-to-end learning frameworks directly map raw sensor inputs to actions using primarily reinforcement~\cite{majumdar2022zson, ramrakhya2022habitat, maksymets2021thda, khandelwal2022simple, ramrakhya2023pirlnav}, leveraging multimodal inputs and large-scale datasets to boost robotic task performance \cite{sun2024prioritized, zhang20233d}. Techniques vary from Transformer-based fusion of image and language to hybrid approaches that improve policy generalization.
Training and evaluating OVON agents require significant variety of scenes and objects, the SD-OVON pipeline, along with its accompanying dataset, can fulfill this need by providing a diverse set of scene variants and object instances for robots to interact with.

\subsection{Prior Datasets and Benchmarks}

Traditional ObjectNav datasets and benchmarks assume the composition of an environment remains constant with objects being immovable fixtures~\cite{yokoyama2024ovon, ramakrishnan2021hm3d, dai2017scannet, Matterport3D}. This setting does not allow a robot to possess prior knowledge of an environment, which contradicts most of our everyday life situations, such as at home or in an office, where objects are frequently rearranged, but prior knowledge of the environment semantics is available. While scene datasets built on synthetic 3D models~\cite{deitke2022ProcTHOR, replica19arxiv} enable object repositioning in dynamic environments, they are constrained to purely synthetic representations that often fail to accurately reflect real-world conditions. Our methodology not only allows for repositioning objects in an environment but also incorporates scans of real-world environments and objects. Compared with popular ObjectNav datasets, such as HM3D~\cite{ramakrishnan2021hm3d} comprising 1000 static scenes, ScanNet~\cite{dai2017scannet} with 1500 static scenes, and MP3D~\cite{Matterport3D} containing 90 static scenes, we expand the number of scene variants from thousands to infinite, by combining static scenes with 0.9k manipulable and movable object models manually inspected and selected from more than 2k Habitat-compatible~\cite{szot2021habitat} object models (see appendix \ref{apx:dataset_details_objects}). The data synthesis and generation pipeline we propose supports dynamic variations of a pre-scanned or artist-created static environment, which facilitates both real-to-sim and sim-to-real transferability.

\section{The SD-OVON Pipeline}
\label{sec:Pipeline}

SD-OVON is a procedural dataset and benchmark generation pipeline. It leverages pretraining VLMs and LLMs to generate photo-realistic scenes, by placing manipulable objects on appropriate receptacles automatically, adhering to everyday life commonsense considering both receptacle and region semantics. For example, \textit{a pillow may be placed on a bed in a bedroom}, and \textit{a cup may appear on a table in a kitchen}. 

\begin{figure}
  \centering
    \includegraphics[width=1.0\linewidth]{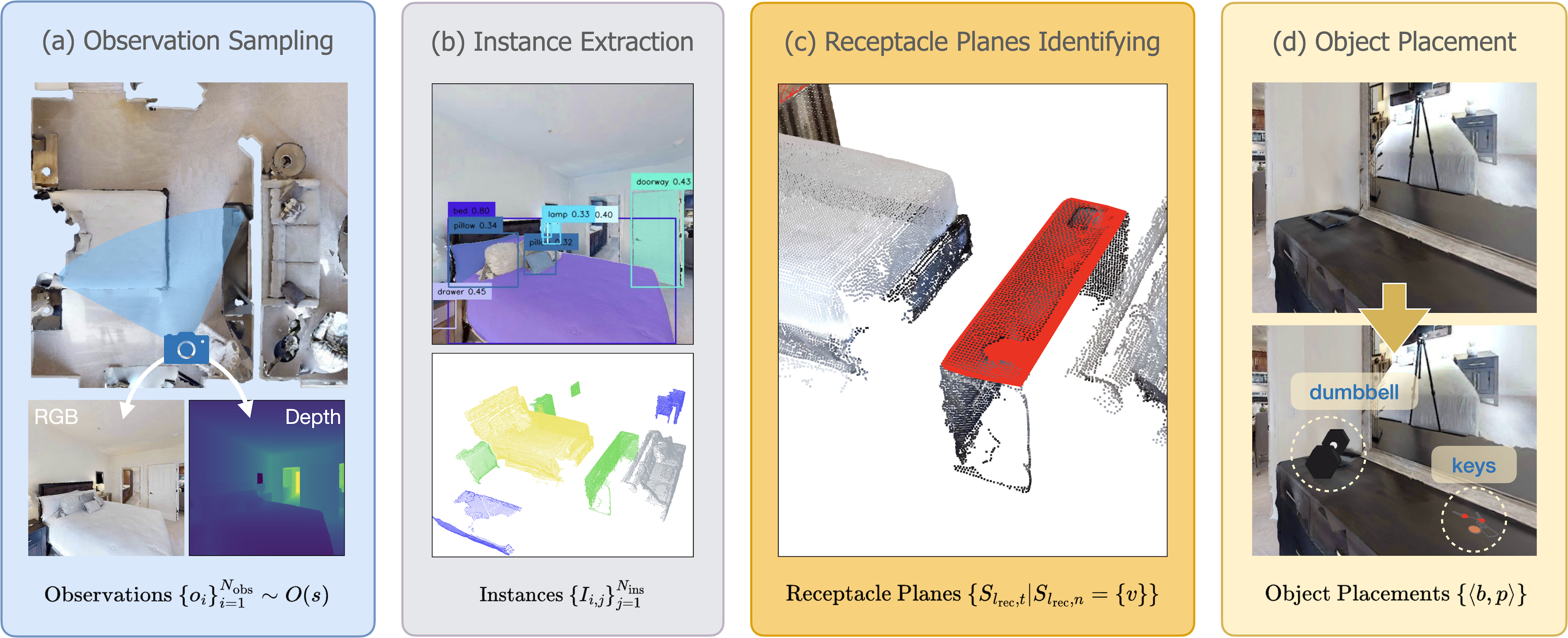}
    \caption{An illustrating of the SD-OVON pipeline. It (a) randomly samples RGB-D observations $o$'s from a scene $s$, (b) extracts and merges open-vocabulary 3D semantic instances $I$'s from the observations, (c) identifies receptacles and available planes $S$'s, and (d) generates scene variants by placing manipulable objects $b$'s at corresponding position $p$'s on appropriate receptacles adhering to daily semantic commonsense.}
    \label{fig:pipeline}
\end{figure}

The SD-OVON pipeline as illustrated in figure~\ref{fig:pipeline} starts by sampling random observations and adopts a Gaussian-based approach to filter out redundant observations while maximizing the 
observation coverage rate (see section \ref{sec:Observation_Sampling}).
Instances are extracted from the observations with open-vocabulary detector, reconstructed in 3D and fused according to spatial and semantic proximity if presented in multiple observation views (see section \ref{sec:Instance_Extraction_and_Fusion}).
Instances within receptacle categories go through a further 3D structure analysis to identify available planes for object placement (see section \ref{sec:Receptable_Planes_Identifying}).
Manipulable objects are then placed on appropriate receptacle planes in corresponding regions, in accordance to the relevance among object categories, receptacle types and region semantics inferred from instance semantics clusters (see section \ref{sec:Semantic_Object_Placement}).
We also integrate our pipeline with the Habitat simulator~\cite{szot2021habitat} to conveniently generate Habitat-compatible ObjectNav task episodes (see section \ref{sec:Task_Episode_Generation}), providing research community with out-of-the-box toolkit in training and evaluating open-vocabulary navigation agents and algorithms.

\subsection{Coverage Random Observations Sampling}\label{sec:Observation_Sampling}

Given a scene $s$ represented in 3D meshes scanned from real-world environments or created by artists, distinct floors are identified by clustering the vertices of meshes. A top-down map $\mathcal{M}_{\text{nav}}(s,i)$ for each floor $i$ is obtained by slicing $s$ horizontally at a specific height ($0.3$m in our implementation) above the floor, which specifies the navigable area of floor $i$ in scene $s$. To ensure observations not placed too close to obstacles, an adjustable erosion is applied to the navigable area on $\mathcal{M}_{\text{nav}}$.
Observations $o \sim O(s)$ are sampled by the observation sampling algorithm $O$ that generates observation points through random sampling on $\mathcal{M}_{\text{nav}}$, followed by a Gaussian-based filter to remove redundant observation points.
In the Gaussian-based filtering, an observation coverage probability layer $\mathcal{M}_{P}$ is initialized on top of $\mathcal{M}_{\text{nav}}$, where the value of each point on $\mathcal{M}_{P}$ indicates the likelihood of being sufficiently observed. On the flip side, the lower the value of a point, the more likely it is to remain as a valid observation point without being filtered out.

Initially, the values of all the points in $\mathcal{M}_{P}$ are set to $0$.
Observation point candidates $p_{\text{cand}} \sim (1 -\mathcal{M}_{P})$ are sampled sequentially. Candidates that are too close to obstacles are filtered out and become invalid. The remaining candidates constitute the set of valid observation points $\{ p_{\text{valid}} \}$.
Once a $p_{\text{valid}}$ is sampled, a set of RGB-D observations $\{ o \}$ is generated by rotating the camera at predefined angular intervals around respectively the yaw and the pitch axes of the camera coordinate system. And a Gaussian distribution $g = \mathcal{N}(p_{\text{valid}}, r_{\max}^{2})$ models the probabilistic observation coverage centered at $p_{\text{valid}}$, given observation range $[r_{\min}, r_{\max}]$ of the camera.
With $g$, the area surrounding $p_{\text{valid}}$ is partitioned into three regions--the \textit{invalid observation region} corresponding to the circular area within radius $r_{\min}$, where the depth camera is too close to obstacles to obtain reliable observations; the \textit{strong observation region} corresponding to the annular region outside the invalid range where $g \ge 0.5$; the \textit{weak observation region} lying beyond the strong range, where $g < 0.5$ and the observation quality degrades due to increased distance.
$\mathcal{M}_{P}$ is updated incrementally with each valid observation point $p_{\text{valid}}$ being sampled and corresponding the Gaussian $g$ overlaid excluding its invalid observation region.
The sampling process stops when either the ratio of area covered by $\mathcal{M}_{P} \geq 0.5$ exceeds a predefined threshold or the number of sampling iterations reaches a predefined threshold.

\subsection{Instance Extraction and Fusion}\label{sec:Instance_Extraction_and_Fusion}

\textbf{Instance extraction.} To extract open-vocabulary semantic instances $I$ from scene $s$, Grounded SAM (G-SAM) \cite{ren2024grounded} is adopted to detect and pixel-wisely segment 2D open-vocabulary instances from each observation $ o = \langle \mathcal{I}, \mathcal{D}, \mathbf{T}_{\text{cam}} \rangle $. Text prompts of instance proposals to G-SAM are automatically generated by open-set image tagging models \cite{huang2023open}. Re-projection of the 3D points reconstructed $ \hat{\mathbf{P}} $ from each depth image $\mathcal{D}$ onto each instance semantics pixel mask $\mathbb{I}$ generated by G-SAM,
\begin{equation}
    \mathbf{U}_{\mathbb{I}}, \mathbf{V}_{\mathbb{I}} = \mathbf{K} \mathbf{T}_{\text{cam}} \hat{\mathbf{P}}
    \label{eq:reprojection}
\end{equation}
with $\mathbf{T}_{\text{cam}}$ being the camera pose of $o$, $\mathbf{K}$ the camera intrinsic matrix, $\mathbf{U}_{\mathbb{I}}$ and $\mathbf{V}_{\mathbb{I}}$ respectively the pixel coordinates on $\mathbb{I}$, further segments these 3D points into point-based geometric semantic representations $ \tilde{\mathbf{P}} $ of the instances presenting in the scene. Each instance $ \mathbf{P} $ can be decomposed into its geometric center $ \bar{p} $ and relative geometry $ \mathbf{P} $
\begin{equation}
\left\langle 
\bar{p}, 
\mathbf{P}
\right\rangle = 
\left\langle 
\frac{1}{N_{\tilde{\mathbf{P}}}}\sum_{p \in \tilde{\mathbf{P}}} p, 
\tilde{\mathbf{P}} - \frac{1}{N_{\tilde{\mathbf{P}}}}\sum_{p \in \tilde{\mathbf{P}}} p
\right\rangle
\label{eq:instance_points_decomposition}
\end{equation}
Associated with object instance semantics label $ l $ of the corresponding instance semantics pixel mask $\mathbb{I}$, an instance geometric representation becomes a semantic geometric instance $ I = \langle \mathbf{P}, \bar{p}, l \rangle $.

\textbf{Instance fusion.} We adapt the object association and fusion algorithms from ConceptGraphs~\cite{gu2024conceptgraphs}, to associate and fuse the semantic geometric instances $I$'s extracted from multiple observation views. For each extracted instance $ I_{i} = \langle \mathbf{P}_{i}, \bar{p}_{i}, l_{i} \rangle $, both semantic similarity $ \varphi_{\text{sem}}(I_i, I_j) $ and geometric proximity $ \varphi_{\text{geo}}(I_i, I_j) $ are evaluated with respect to each fused instance $I_{j}$. 
$\varphi_{\text{sem}}$ is given by the overlap between point clouds $ (\mathbf{P}_{i} + \mathbf{1} p_{i}^{\mathsf{T}}) $ and $ (\mathbf{P}_{j} + \mathbf{1} p_{j}^{\mathsf{T}}) $, and $ \varphi_{\text{geo}} $ is given by the normalized cosine distance between the semantics $l_{i}$ and $l_{j}$. The overall similarity
\begin{equation}
    \varphi(I_i, I_j) = k_{\text{sem}}\varphi_{\text{sem}}(I_i, I_j) + k_{\text{geo}}\varphi_{\text{geo}}(I_i, I_j)
\end{equation}
is weighted by $k_{\text{sem}}$ and $k_{\text{geo}}$, and identifies the best match for instance $I_{i}$ to a fused instance with the highest overall similarity score over certain threshold $\varphi_{\min}$. If no suitable match is found with all $\varphi < \varphi_{\min}$, a new fused instance is instantiated with $I_{i}$. Upon successful association between $I_{i}$ and $I_{j}$, the $(\mathbf{P}_{i} + \mathbf{1} p_{i}^{\mathsf{T}})$ and $(\mathbf{P}_{j} +\mathbf{1} p_{j}^{\mathsf{T}})$ are merged and downsampling follows to remove redundant points, ensuring the representation of each fused instance is progressively refined over time. And refined point clouds are decomposed following equation \ref{eq:instance_points_decomposition} to form the updated fused instance $I_{j}'$.

\textbf{Fusion error correction.} Due to imperfect label detection with G-SAM, partial instances detected from multiple views of an object may be associated with different labels, and therefore become different fused instances.
To address this inconsistency and enhance the accuracy of instance fusion, we propose an error correction mechanism that mitigates errors introduced at the instance fusion stage. A pairwise comparison identifies fused instances $I_{i}$ and $I_{j}$ that exhibit significant spatial overlap with $\text{IoU}(I_{i}, I_{j}) \geq \text{IoU}_{\min}$. For such fused instance pairs, the averaged G-SAM detection confidence scores of the top-$k$ partial instances constituting respectively $I_{i}$ and $I_{j}$ are compared, where $k$ is the minimum number of constituting partial instances of $I_{i}$ and $I_{j}$. The fused instance with lower average confidence score is considered less reliable and subsequently discarded.

\subsection{Receptacle Planes Identifying}\label{sec:Receptable_Planes_Identifying}

Receptacles, where manipulatable objects may be placed, are identified from the fused instances given in section \ref{sec:Instance_Extraction_and_Fusion} by the list of receptacle labels $ L_{\text{rec}} = \{ l_{\text{rec}} \} $, such as \textit{table}, \textit{shelf} and \textit{bed}. $ L_{\text{rec}} $ can be identified through LLMs (see appendix \ref{apx:llm_prompt_identifying_receptacles}) or prespecified. To detect and extract available planes for object placement of a receptacle, we propose an expectation-maximization (EM) plane detection algorithm (algorithm \ref{alg:em_plane_detection}).
\begin{algorithm}[ht]
\caption{EM Plane Detection}
\label{alg:em_plane_detection}
\KwIn{receptacle $\mathbf{P}$, plane thickness $\epsilon_{t}$, plane detection ratio $\rho_{\min}$, convergence tolerance $\epsilon_{c}$}
\KwOut{List of receptacle planes $L_{\text{RP}}$}
\SetKw{KwStep}{step}
\SetKw{KwBreak}{break}
\SetKw{KwTrue}{true}
\SetKwFunction{KwFuncConvexProjectedHull}{ConvexProjectedHull}
\BlankLine
Initialize $L_{\text{RP}} \gets \{\}$\;
\For{$h_{i} \gets h_{\min}(\mathbf{P})$ \KwTo $h_{\max}(\mathbf{P})$ \KwStep $0.5\epsilon_{t}$}{
    Collect points $\mathbf{P}_{i} = \{ p | h_{i} \leq h(p) \leq h_{i} + \epsilon_{t} \}$ of the potential plane\;
    \If{$|\mathbf{P}_{i}| \geq \rho_{\min}$}{
        Reset $\mathbf{P}' \gets \{\}$\;
        \While{\KwTrue}{
            Fit a plane $S_{i}$ and find its normal vector $v_{i}$ with $\mathbf{P}_{i}$\;
            Collect points $\mathbf{P}'$ between $\pm\epsilon_{t}$ along $v_{i}$ from plane $S_{i}$\;
            \If{$| \mathbf{P}' \cap \mathbf{P}_{i} | \geq \epsilon_{c} |\mathbf{P}_{i}| $}{
                \KwBreak\;
            }
            $\mathbf{P}_{i} \gets \mathbf{P}'$\;
        }
        $L_{RP} \gets L_{RP} + \{ \KwFuncConvexProjectedHull(\mathbf{P}_{i},S_{i})\}$
    }
}
\textbf{End For}
\end{algorithm}
In \texttt{ConvexProjectedHull}, the filtered inlier points are projected onto the final fitted plane $S_{i}$, the convex hull for these projected points is computed and represented by its vertices in counter-clockwise order around the Z-axis. This approach yields a set of robust, accurately fitted planes suitable for object placement within the scene.

\subsection{Region-receptacle Semantics-aware Object Placement}\label{sec:Semantic_Object_Placement}

\textbf{Probabilistic semantics relating.} We adapt a sliding window approach~\cite{qiu2024open} to scan through $s$ by projecting fused instances $I$'s to top-down perspective and generate a semantic map $\mathcal{M}$ with dense open-vocabulary regional semantics by leveraging the reasoning capabilities of state-of-the-art LLMs~\cite{grattafiori2024llama} (see appendices \ref{apx:llm_prompt_region_proposals}, \ref{apx:llm_prompt_receptacle_relevance} and \ref{apx:llm_prompt_region_relevance}). The circular sliding window with radius $ r $ moves by step size $ \Delta d $. At echo step $(x_{w}, y_{w})$, the area within the sliding window
\begin{equation}
    W(x_{w}, y_{w}) = \left\{ (x,y) | (x - x_{w}\Delta d)^2 + (y - y_{w}\Delta d)^2 \leq r^{2} \right\}
\end{equation}
selects a set of object instances falling inside, whose labels after repeated terms removed effectively describe the objects presenting within this area. We then leverage the zero-shot abstract reasoning capability of LLMs \cite{grattafiori2024llama} to come up with a list of region proposals, depicting the abstract region semantics of the area. After a full sweep, a zero-shot dense prediction of region semantics $ R(x,y) $ over the complete map is generated, constituting the semantic map $\mathcal{M}$ that represents the regional semantics of the entire floor, which region labels such as \textit{kitchen}, \textit{bedroom}, \textit{living room}, etc.

Normalized semantic relevance scores $ \mathbb{P}(l_{\text{obj}}|l_{\text{rgn}}) $ between manipulable object categories $l_{\text{obj}}$ and regions $l_{\text{rgn}}$ from $\mathcal{M}$, and $ \mathbb{P}(l_{\text{obj}}|l_{\text{rec}}) $ between manipulable object categories $l_{\text{obj}}$ and receptacle categories $l_{\text{rec}}$ are further given by LLMs \cite{grattafiori2024llama}. The joint object-region-receptacle relevance 
\begin{equation}
    \mathbb{P}(l_{\text{obj}}|l_{\text{rgn}},l_{\text{rec}}) = \mathbb{P}(l_{\text{obj}}|l_{\text{rgn}})\mathbb{P}(l_{\text{obj}}|l_{\text{rgn}})
    \label{eq:rro_joint_relevance}
\end{equation}
is then derived to determine the appearance probabilities of manipulable objects across receptacles and regions during object placement.

\textbf{Semantics-aware and Physics-constrained Object Placement.} For each receptacle $l_{\text{rec}}$ at each region $l_{\text{rgn}}$, we sample $m$ repeatable categories $\{ l_{\text{obj},i} \}_{i=1}^{m}$ for manipulable objects. The exact object instance $ I_{\text{obj},i} $ under category $ l_{\text{obj},i} $ is randomly selected to be place on an available plane $S$ of $l_{\text{rec}}$. For the exact coordinate to spawn $ l_{\text{obj},i} $, we adopt a uniform position sampling algorithm to sample candidate positions on a convex polygonal plane (see appendix \ref{apx:uniform_sampling_in_convex_polygons}) of $l_{\text{rec}}$ identified from section \ref{sec:Receptable_Planes_Identifying}. To prevent collisions between object instances, a minimum distance constraint between each object instance pairs is enforced to avoid object instances from overlapping. $ l_{\text{obj},i} $ is then spawned at a certain height ($0.3$m in our implementation) above the plane, and randomly rotated along the $X$, $Y$, and $Z$ axes to introduce variance of the scene. Physics simulation follows after all object instances are spawned, to allow all objects falls freely onto receptacle planes until they reach a stable state. This process ensures objects settle in physically feasible positions and enhances the scene fidelity.

\subsection{ObjectNav Task Episode Generation}\label{sec:Task_Episode_Generation}

We also incorporate our pipeline with the Habitat simulator~\cite{szot2021habitat}, allowing automatic generation of Habitat-compatible ObjectNav task episodes. For each generated scene variant with well-placed objects, goal viewpoints as ObjectNav success criteria are generated by sampling navigable areas surrounding each object. Goal viewpoints are then doubled-checked to ensure the goal object can be seen from them, and those without valid sight of the goal object will be filter out. If an object can not be seen from any viewpoint, it will be considered as invalid and therefore removed from the list of goals. A configurable number of episodes are then generated for each object category as part of the task episode dataset. An object instance under the object category is randomly chosen within the scene, when an object category is selected as the navigation goal in an episode. The total number of episodes generated in an ObjectNav task dataset is given by
\begin{equation}
    N = k_{s} \times k_{v} \times n_{c} \times k_{e}
\end{equation}
with configurable $k_{s}$, $k_{v}$ and $k_{e}$ being the number of static scenes used for data generation, the number of variants to generate per scene, the number of episodes to generate per object category in a scene variant, respectively. And $n_{c}$ is the average number of valid object categories in the scene variants.

\section{Datasets}\label{sec:Dataset}

\begin{table}
\centering
\caption{Statistics of data sources and the \textit{SD-OVON-Objects} datasets.}
\label{tab:datasets}
\begin{tabular}{c c c c c}
    \toprule
     & Dataset & \# of Instances & \# of Categories & Type \\
    \midrule
    Source & YCB~\cite{calli2015ycb} & 77 & 56 & Scanned \\
     & AI2-THOR~\cite{kolve2017ai2} & 469 & 36 & Artist-created \\
     & ABO~\cite{collins2022abo} & 195 & 20 & Artist-created \\
     & HSSD~\cite{khanna2023hssd} & 1003 & 24 & Artist-created \\
    \midrule
    Ours & SD-OVON-Objects-Raw & 1744 & 110 & Hybrid \\
     & SD-OVON-Objects & 889 & 73 & Hybrid \\
    \bottomrule
\end{tabular}
\end{table}

\begin{table}
\centering
\caption{Statistics of the \textit{SD-OVON} ObjectNav task datasets.}
\label{tab:subdatasets_eval}
\begin{tabular}{c c c c c c}
    \toprule
    Dataset & \# of Categories & Scenes & Scene Variants & \# of Episodes \\
    \midrule
    \textit{SD-OVON-3k} & 73 & 8 & 363 & 2897 \\
    \textit{SD-OVON-10k} & 73 & 20 & 1127 & 10629 \\
    \bottomrule
\end{tabular}
\end{table}

\textbf{The SD-OVON-Objects datasets.}
To equip the SD-OVON pipeline with a ready-to-use object dataset for generating scene variants and compatible with simulation in Habitat~\cite{szot2021habitat}, we curate hybrid object model data consisting of both photo-realistic scans of real-world objects~\cite{calli2015ycb} and artist-created 3D models~\cite{kolve2017ai2, collins2022abo, khanna2023hssd} of daily objects, forming the \textit{SD-OVON-Objects-Raw} dataset. We also manually inspect each object model and select those with high visual fidelity to form the \textit{SD-OVON-Objects} dataset. Details about the datasets are listed on table \ref{tab:datasets} and can be found in appendix \ref{apx:dataset_details_objects}.

\textbf{The SD-OVON ObjectNav task datasets.}
Leveraging the SD-OVON pipeline and the SD-OVON-Objects dataset and photo-realistic scenes from \cite{ramakrishnan2021hm3d, dai2017scannet, Matterport3D}, we pre-generate and 13k ObjectNav task episodes along with 1.5k scene variants, comprised in the \text{SD-OVON-3k} and the \text{SD-OVON-10k} ObjectNav task datasets as shown on table \ref{tab:datasets} (see appendix \ref{apx:dataset_details_task} for more details), for ready-to-use training and evaluation of ObjectNav and OVON agents. It is noteworthy that SD-OVON-3k focuses more on manipulatable and movable daily objects, like \textit{books}, \textit{cups}, and \textit{stuffed toys}, which may be repositioned irregularly in dynamic environments in daily life, instead of static furniture, such as \textit{bed} and \textit{sofa}, in traditional ObjectNav task episodes datasets~\cite{yokoyama2024ovon, Matterport3D, khanna2023hssd, szot2021habitat}. The dynamic scene variants generated are compatible for the open-vocabulary mobile manipulation tasks~\cite{yenamandra2024homerobotopenvocabularymobilemanipulation} as well.

\section{Baselines and Experiment}\label{sec:Experiment}

\subsection{Baselines}

To validate the effectiveness of the SD-OVON pipeline and the accompanied datasets, we propose two baselines, the Random Receptacle Navigation A* and the Semantic Navigation A*, and evaluate them along with state-of-the-art OVON baselines~\cite{yokoyama2024vlfm, yu2023l3mvn}.

\textbf{Random A*.} Random Receptacle Navigation A* (Random A*) preserves prior knowledge about an environment, which is a common setting for daily situations. During the initial visit to a new environment, it explores the environment using frontier-based methods~\cite{sun2020frontier, cao2022autonomous} and obtains comprehensive observations of the environment. It then identifies receptacles in the environment following the similar approach introduced in section \ref{sec:Receptable_Planes_Identifying}. At each following episode, the order of receptacles to visit are randomly shuffled, and candidate navigation points are generated surrounding each receptacle in the shuffled order. GroundingDINO~\cite{liu2023grounding} is used for open-vocabulary object detection at each navigation point, and the agent switches to point goal navigation mode after the target object is detected. (See appendix \ref{apx:random_astar_details} for implementation details.)

\textbf{Semantic A*.} Semantic Navigation A* (Semantic A*) is follows similar procedure to the Random A*. However, it not only extract receptacle planes but also build region-receptacle semantic awareness of the environment during the initial exploration stage, following the instance extraction, fusion and the region-receptacle semantics inference approaches as introduced from section \ref{sec:Instance_Extraction_and_Fusion} to section \ref{sec:Semantic_Object_Placement}. Furthermore, Semantic A* prioritizes the receptacles to visit by decreasing order of relevance scores of the target object to region-receptacle pairs following equation \ref{eq:rro_joint_relevance}, rather than randomly explores the receptacles in a scene. (See appendix \ref{apx:semantic_astar_details} for more details.)

\textbf{VLFM.} Vision-Language Frontier Maps (VLFM)~\cite{yokoyama2024vlfm} is a training-free modular approach that achieves state-of-the-art performance on various ObjectNav benchmarks. Besides building occupancy map from the depth sensing and odometry, VLFM leverages VLMs to build semantic relevance between the goal object and the explored regions in the map. It prioritizes its search for the goal object in accord to the region semantic relevance. By integrating GroundingDINO~\cite{ren2024groundingdino15}, VLFM also exhibits open-vocabulary object detection and navigation capabilities.

\textbf{L3MVN.} Leveraging Large Language Models for Visual Target Navigation (L3MVN)~\cite{yu2023l3mvn} is another training-free modular approach. L3MVN builds a semantic map from the obtained observations, and extracts the frontier map from the explored map and the obstacle map. It employs LLMs to infer the semantic relevance between the goal object and the observations, and then select a long-term goal to approach from all the frontiers based on the relevance score.

\subsection{Evaluation and Analysis}\label{sec:eval_and_analysis}

\begin{table}
\centering
\caption{Evaluation results of the baselines on the SD-OVON-3k dataset.}
\label{tab:results}
\begin{tabular}{c c c c c}
    \toprule
    Baseline & Success Rate($\uparrow$) & SPL($\uparrow$) & Soft SPL($\uparrow$) & Dist2Goal($\downarrow$) \\
    \midrule
    VLFM & 0.0204 & 0.0059 & 0.0554 & 7.5077 \\
    L3MVN & 0.0159 & 0.0025 & - & 7.5604 \\
    Random A* & 0.0974 & 0.0307 & 0.1057 & \textbf{5.4072} \\
    Semantic A* & \textbf{0.1417} & \textbf{0.0647} & \textbf{0.1539} & 5.5671 \\
    \bottomrule
\end{tabular}
\end{table}

\begin{figure}
    \centering
    \includegraphics[width=0.32\linewidth]{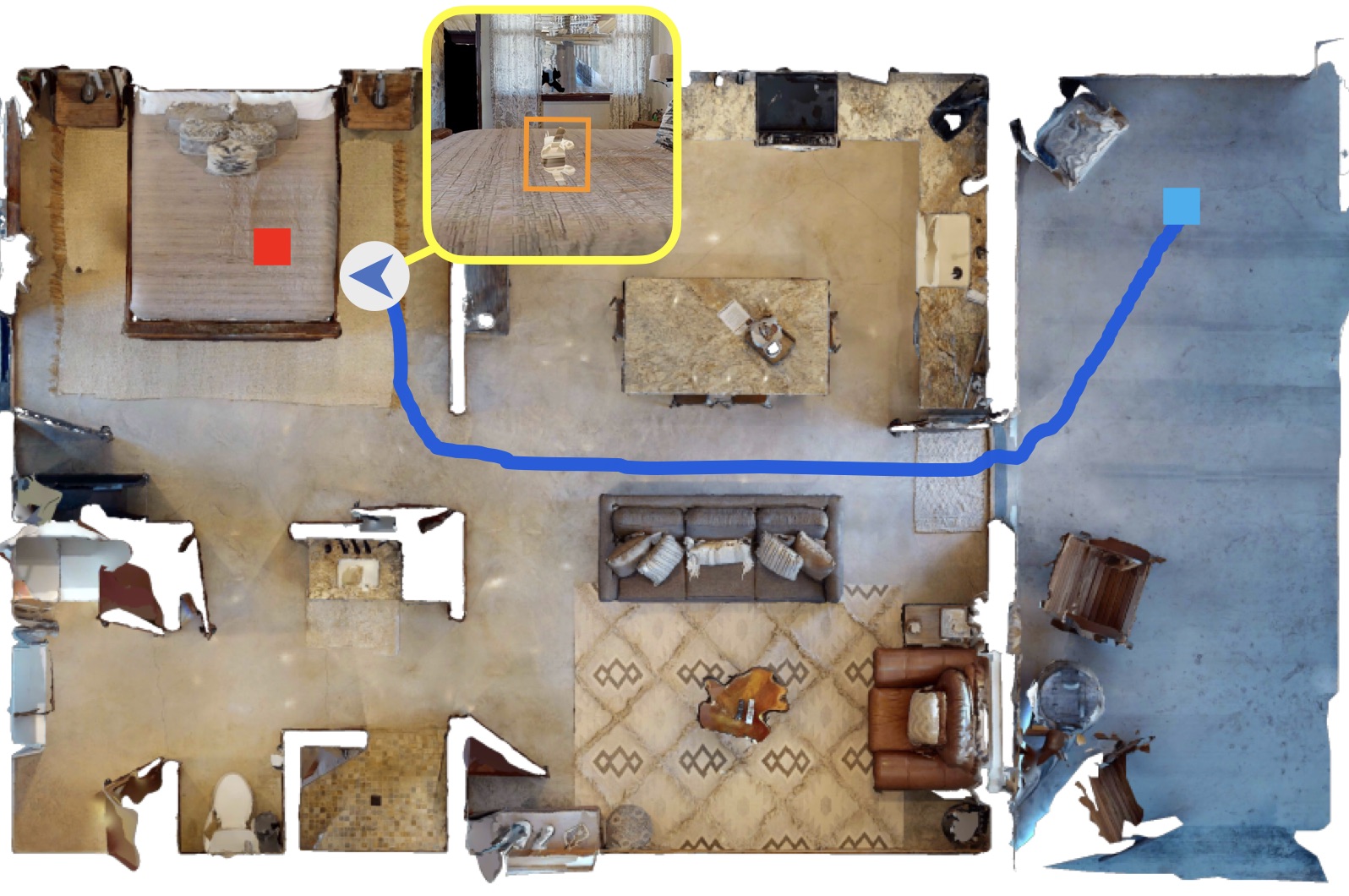}
    \includegraphics[width=0.32\linewidth]{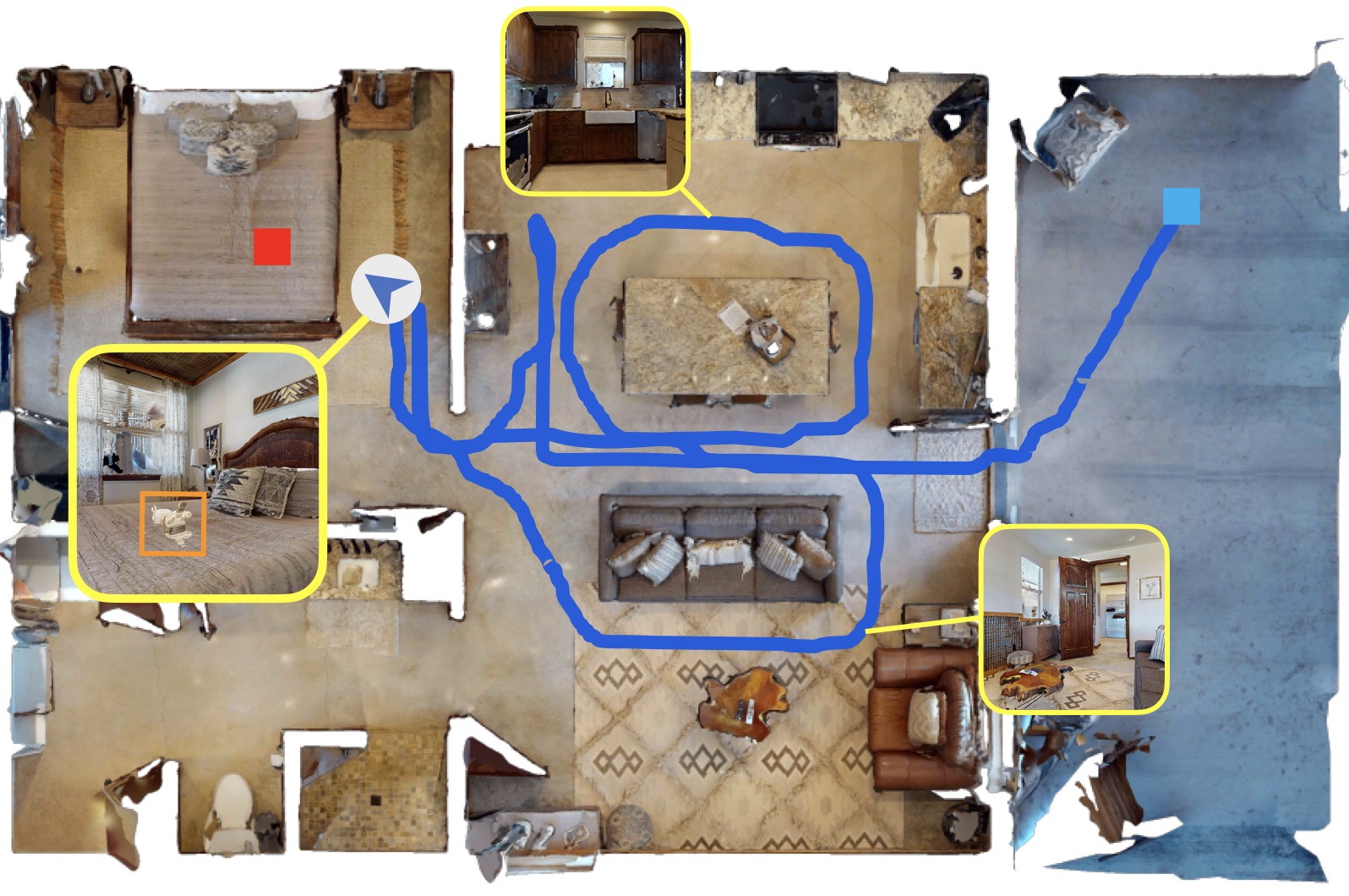}
    \includegraphics[width=0.32\linewidth]{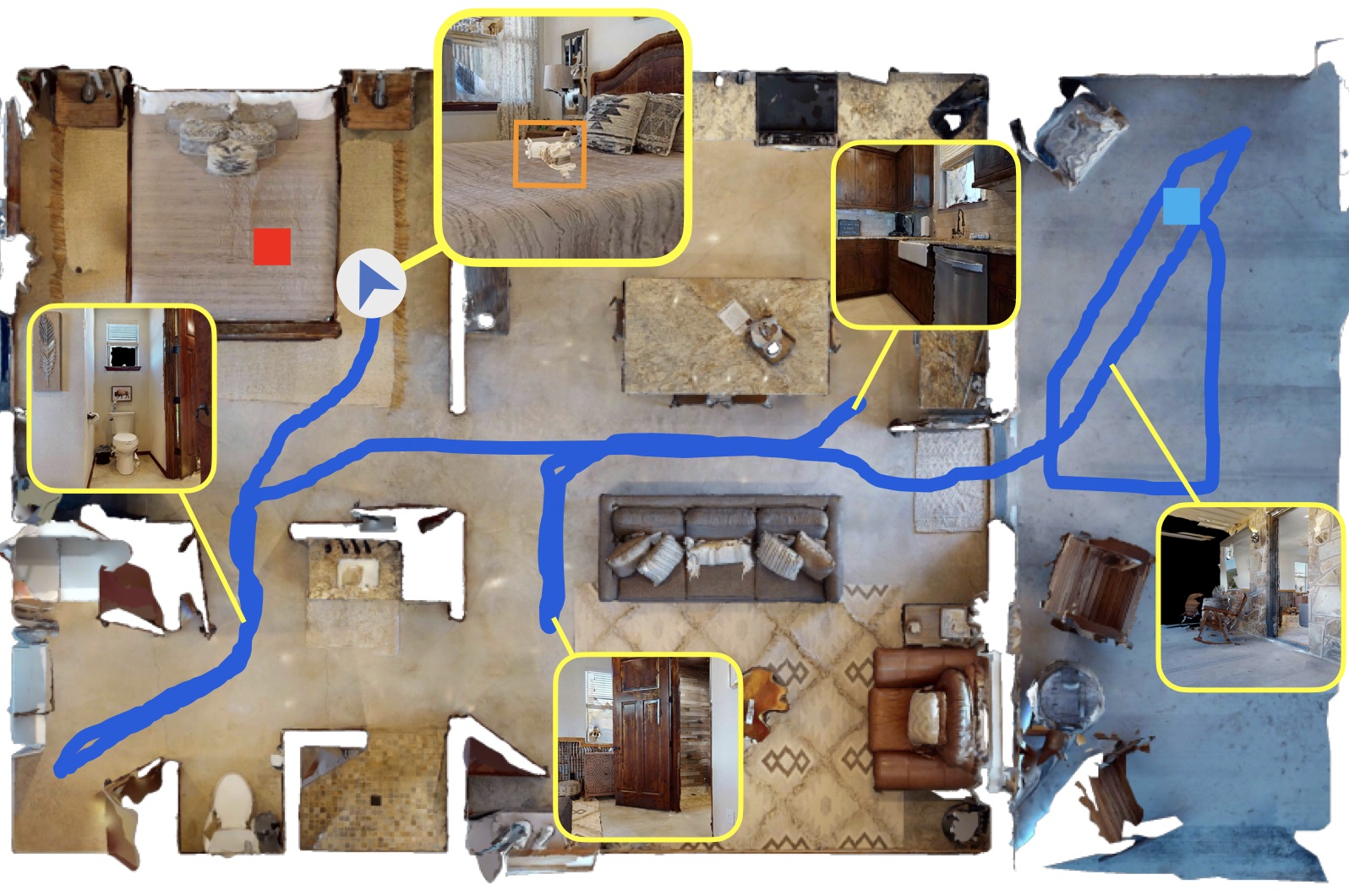}
    \caption{Example trajectories of successful navigation with \textbf{Semantic A*} (left), \textbf{Random A*} (middle) and \textbf{VLFM}~\cite{yokoyama2024vlfm} (right) respectively for an ObjectNav task episode from SD-OVON-3k.}
    \label{fig:successful_examples}
\end{figure}

We evaluate Random A*, Semantic A*, VLFM~\cite{yokoyama2024vlfm} and L3MVN~\cite{yu2023l3mvn} on the SD-OVON-3k dataset, and report the average success rate (SR), the average success weighted by inverse path length (SPL)~\cite{anderson2018evaluation}, Soft SPL and the average fianl distance to goal (Dist2Goal) for each baseline. SPL scores both the success of navigation and the efficiency of an agent.

Evaluation results, as listed in \ref{tab:results}, reveals that state-of-the-art OVON baselines \cite{yokoyama2024vlfm, yu2023l3mvn} meet challenges in solving the open-vocabulary object navigation task with dynamic environmental change settings that in fact adhere to our daily situations, such as at home or at work. And both our proposed baselines, the Random A* and the Semantic A*, beat state-of-the-art OVON baselines on the SD-OVON-3k dataset. And the Semantic A* agent with both region and receptacle semantic awareness of the visited environments outperforms other baselines, demonstrating the significance of the semantics-aware dynamic environment setting we propose as a novel OVON task, and the effectiveness of our proposed SD-OVON pipeline and its accompanied datasets. Visualization of example trajectories of successful navigation, as shown in figure \ref{fig:successful_examples}, straightforwardly illustrates the advantage of possessing environmental context in familiar scenes, where Semantic A* agent goes directly towards the \textit{stuff toy} on the bed in the bedroom, while Random A* and VLFM~\cite{yokoyama2024vlfm} agents require significant exploration before reaching the goal. Further investigation and analysis are presented in section \ref{sec:ablation}.

\subsection{Ablation Studies}\label{sec:ablation}

\textbf{Ablation study on detector performance.}
Further investigation into each approach reveals that the performance of detectors significantly affects the overall performance of OVON agents. As shown in table \ref{tab:ablation_detector}, with the detector of each baseline replaced by a ground-truth semantics detector (GT), the overall performance of each agent increases significantly. It suggests that a better detector can boost the overall performance of an OVON agent to a great extend. However, the Semantic A* agent still beats other baselines, drawing the same conclusion as in section \ref{sec:eval_and_analysis} about the importance of semantics accessibility to familiar environments, the significance of the dynamic environment OVON setting we bring up that adheres to everyday situations and semantics, and the effectiveness of the SD-OVON pipeline and the two baselines we propose.

\begin{table}[t]
\centering
\caption{Ablation study on detector performance of the baselines on SD-OVON-3k.}
\label{tab:ablation_detector}
\begin{tabular}{c c c c c}
    \toprule
    Baseline & Success Rate($\uparrow$) & SPL($\uparrow$) & Soft SPL($\uparrow$) & Dist2Goal($\downarrow$) \\
    \midrule
    VLFM & 0.0204 & 0.0059 & 0.0554 & 7.5077 \\
    VLFM GT & \textbf{0.3935} & \textbf{0.1557} & \textbf{0.1840} & \textbf{4.8796} \\
    \midrule
    L3MVN & 0.0159 & 0.0025 & - & 7.5604 \\
    L3MVN GT & \textbf{0.5875} & \textbf{0.2629} & - & \textbf{3.3319} \\
    \midrule
    Random A* & 0.0974 & 0.0307 & 0.1057 & 5.4072 \\
    Random A* GT w/fa & \textbf{0.7503} & \textbf{0.3494} & \textbf{0.3931} & \textbf{1.5437} \\
    \midrule
    Semantic A* & 0.1417 & 0.0647 & 0.1539 & 5.5671 \\
    Semantic A* GT & \textbf{0.8081} & \textbf{0.4270} & \textbf{0.4700} & \textbf{1.0989} \\
    \bottomrule
\end{tabular}
\end{table}

We also evaluated the categorical performances of different baselines on the SD-OVON-3k dataset. The top-5 categories on success rates with at least 2 successful episodes, to avoid incidental success, of different baselines are listed in table \ref{tab:ablation_categorical_performance}. Random A* and Semantic A* demonstrate much higher success rates than both VLFM~\cite{yokoyama2024vlfm} and L3MVN~\cite{yu2023l3mvn} across the top-5 categories. It is noticed that VLFM and L3MVN favors objects with larger sizes that can be detected more easily, such as \textit{boxes}, \textit{cushions} and \textit{folders}, inferring that the detector plays a crucial role in the OVON task especially encountered with small daily objects. On the other hand, however, Semantic A* and Random A* are capable of finding small daily items like \textit{spoons}, \textit{mugs} and \textit{cans} with relatively high success rates, even if these two baselines employs the same detector~\cite{ren2024groundingdino15} as VLFM. It again suggests the significance and effectiveness of exploiting environment context and semantics of familiar environments, the same strategy people would employ in their daily life.

\begin{table}[t]
\centering
\caption{Ablation study on categorical performances of the baselines on SD-OVON-3k. Top-5 categories on success rates with at least 2 successful episodes of different baselines are listed. \textbf{Cat.} stands for object category and \textbf{SR} for categorical success rate.}
\label{tab:ablation_categorical_performance}
\begin{tabular}{ccccc}
    \toprule
    Top-k / Cat. (SR) & VLFM~\cite{yokoyama2024vlfm} & L3MVN~\cite{yu2023l3mvn} & Random A* & Semantic A* \\
    \midrule
    Top-1 & box         & folder      & soap dispenser  & spoon \\
          & (0.1389)    & (0.0588)    & (0.4286)        & (0.6000) \\
    \midrule
    Top-2 & cushion     & jug         & hammer          & hammer \\
          & (0.1200)    & (0.0455)    & (0.2941)        & (0.4444) \\
    \midrule
    Top-3 & power drill & gelatin box & dumbbell        & cushion \\
          & (0.0556)    & (0.0377)    & (0.2750)        & (0.4021) \\
    \midrule
    Top-4 & handbag     & vase        & spoon           & laptop \\
          & (0.0541)    & (0.0377)    & (0.2222)        & (0.3714) \\
    \midrule
    Top-5 & stuffed toy & cup         & potted meat can & mug \\
          & (0.0370)    & (0.0330)    & (0.2143)        & (0.3684) \\
    \bottomrule
\end{tabular}
\end{table}

\textbf{Ablation study on final point goal navigation strategy.}
We investigate the effectiveness of the final point goal navigation strategy of the two proposed baselines, the Random A* and the Semantic A*, on the SD-OVON-3k dataset. To avoid being affected by the detector performance, we compare the overall performance of both baselines with the ground-truth semantics detector (GT) between the settings with and without the final point goal navigation strategy (FN). Results of the ablation experiment are show on table \ref{tab:ablation_final_approach}. The results reveals that the final point goal navigation strategy has significant impact on the performance of the baseline agents. Both the Semantic A* and the Random A* agents doubled their performance when the final point goal navigation strategy is applied, compared to the ones without the strategy.

\begin{table}[t]
\centering
\caption{Ablation study of final point goal navigation strategy (FN) of Random A* and Semantic A*.}
\label{tab:ablation_final_approach}
\begin{tabular}{c c c c c}
    \toprule
    Baseline & Success Rate($\uparrow$) & SPL($\uparrow$) & Soft SPL($\uparrow$) & Dist2Goal($\downarrow$) \\
    \midrule
    Random A* GT w/o FN & 0.3041 & 0.1471 & 0.3748 & 2.1105 \\
    Random A* GT & \textbf{0.7503} & \textbf{0.3494} & \textbf{0.3931} & \textbf{1.5437} \\
    \midrule
    Semantic A* GT w/o FN & 0.3752 & 0.2111 & \textbf{0.4742} & 1.5230 \\
    Semantic A* GT & \textbf{0.8081} & \textbf{0.4270} & 0.4700 & \textbf{1.0989} \\
    \bottomrule
\end{tabular}
\end{table}

\section{Conclusion and Discussion}
\label{sec:Conclusion}

In this work, we propose the Semantics-aware Dataset and Benchmark Generation Pipeline for Open-vocabulary Object Navigation in Dynamic Scenes (SD-OVON). SD-OVON is a procedural pipeline that synthesizes and generates infinite photo-realistic scene variants and ObjectNav task episodes from static scans of and/or artist-created static scenes and object models, adhering to everyday situations and social norms. Our work fills the gap between the traditional open-vocabulary object navigation task and dynamic daily scenarios in the real world, where people often navigate in familiar environments but the objects in these environments are subject to frequent rearrangement in daily basis, rather than explore new environments every time. 
The extension of the open-vocabulary object navigation problem, from considering only large objects and static furniture in most previous research to involving small movable and manipulatable daily items in this work, bridges the gap between open-vocabulary object navigation algorithms and their practical applications on physical robotics and embodied intelligence systems designed for dynamic real-world situations. It brings us another step closer to the massive deployment of generalist service robots in our daily life to serve people, especially the elderly.

Accompanied with the SD-OVON pipeline, we also offer two pre-generated ObjectNav task datasets with respectively 3k and 10k Habitat-compatible ObjectNav task episodes derived from 0.9k hybrid object models we collect and manually inspect. The task and the object datasets, along with the scene dataset consists of over 2.5k static scenes, are available for out-of-the-box training and evaluation of ObjectNav and OVON agents in the Habitat~\cite{szot2021habitat} simulation environment. Evaluation of the Semantic A*, the Random A* and state-of-the-art baselines on the ObjectNav task dataset we pre-generate demonstrates the effectiveness of our propose pipeline and datasets on the OVON task with dynamic environment settings adhering to daily situations and semantics.

Though our proposed pipeline allows semantics-aware placement of objects on receptacle surfaces, the it does not considered placing objects inside articulated receptacles in our current implementation, such as drawers and cabinets with doors, or rearranging furniture layout. Extending the SD-OVON pipeline towards more flexible forms of object placement facilitates the training of OVON agents with better adaptability to real-world environments and more comprehensive evaluation of such agents. The SD-OVON pipeline depends on preexisting object models, which significantly limits the generality of the pipeline. Latest advance in generative technologies, especially 3D assets generation, can be incorporated into the pipeline to automatically generate movable and manipulatable object models for placement on/in receptacles. The consideration of dynamic scenes and accessible prior knowledge of the environment is a new task setting of the open-vocabulary object navigation task that effectively reflects our everyday situations. Therefore, we propose two baselines, the Random A* and the Semantic A*, as an initial exploration on tackling such a task. However, more exploration and effort from the research community are necessary to completely resolve this problem. Besides, the movable and manipulatable nature of the object models placed in the generated scene variants also allows convenient extension of the scene variants for the open-vocabulary mobile manipulation task.

{
    \small
    \bibliographystyle{unsrtnat}
    \bibliography{nips/refs}
}

\clearpage

\appendix

\section{Dataset Details}\label{apx:dataset_details}

\subsection{The SD-OVON-Objects Datasets}\label{apx:dataset_details_objects}

We curate object model data with both photo-realistic scans of real-world objects and artist-created 3D models of daily objects. The primary dataset, SD-OVON-Objects on table \ref{tab:datasets} (or \textit{SD-OVON-Objects-0.9k-MI} on table \ref{tab:datasets_extended}), consists of object models manually inspected and selected from SD-OVON-Objects-Raw on table \ref{tab:datasets} (or \textit{SD-OVON-Objects-1.7k} on table \ref{tab:datasets_extended}). SD-OVON-Objects-1.7k consists of all the object models from YCB~\cite{calli2015ycb}, the AI2-THOR~\cite{kolve2017ai2}, the ABO~\cite{collins2022abo} and the HSSD~\cite{khanna2023hssd} datasets. We also extend SD-OVON-Objects-1.7k, by merging it with the object models from the Google Scanned Objects (GSO) dataset~\cite{downs2022google}, to \textit{SD-OVON-Objects-2k}. The three object datasets we provided are compatible to the Habitat~\cite{szot2021habitat} environment for immediate usage.

\begin{table}[ht]
\centering
\caption{Statistics of the extended data sources and the extended \textit{SD-OVON-Objects} datasets.}
\label{tab:datasets_extended}
\begin{tabular}{c c c c}
    \toprule
    Dataset & \# of Instances & \# of Categories & Type \\
    \midrule
    YCB~\cite{calli2015ycb} & 77 & 56 & Scanned \\
    AI2-THOR~\cite{kolve2017ai2} & 469 & 36 & Artist-created \\
    ABO~\cite{collins2022abo} & 195 & 20 & Artist-created \\
    HSSD~\cite{khanna2023hssd} & 1003 & 24 & Artist-created \\
    GSO~\cite{downs2022google} & 342 & 58 & Scanned \\
    \midrule
    SD-OVON-Objects-2k & 2086 & 153 & Hybrid \\
    SD-OVON-Objects-1.7k & 1744 & 110 & Hybrid \\
    SD-OVON-Objects-0.9k-MI & 889 & 73 & Hybrid \\
    \bottomrule
\end{tabular}
\end{table}

\subsection{SD-OVON-Scenes}\label{apx:dataset_details_scenes}

We curate static scenes consisting of primarily photo-realistic scans of real-world environments in the \textit{SD-OVON-Scenes} dataset. The SD-OVON-Scenes dataset includes over 2.5k static scenes, and relevant details are listed on table \ref{tab:scenes_datasets}.

\begin{table}[ht]
\centering
\caption{Statistics of the data sources and the \textit{SD-OVON-Scenes} dataset.}
\label{tab:scenes_datasets}
\begin{tabular}{c c c}
    \toprule
    Dataset & \# of Static Scenes & Type \\
    \midrule
    HM3D~\cite{ramakrishnan2021hm3d} & 1000 & Scanned \\
    MP3D~\cite{Matterport3D} & 90 & Scanned \\
    ScanNet~\cite{dai2017scannet} & 1613 & Scanned \\
    \midrule
    SD-OVON-Scenes & 2703 & Scanned \\
    \bottomrule
\end{tabular}
\end{table}

\subsection{The SD-OVON ObjectNav Task Datasets}\label{apx:dataset_details_task}

\begin{figure}[ht]
    \centering
    \includegraphics[width=1.0\linewidth]{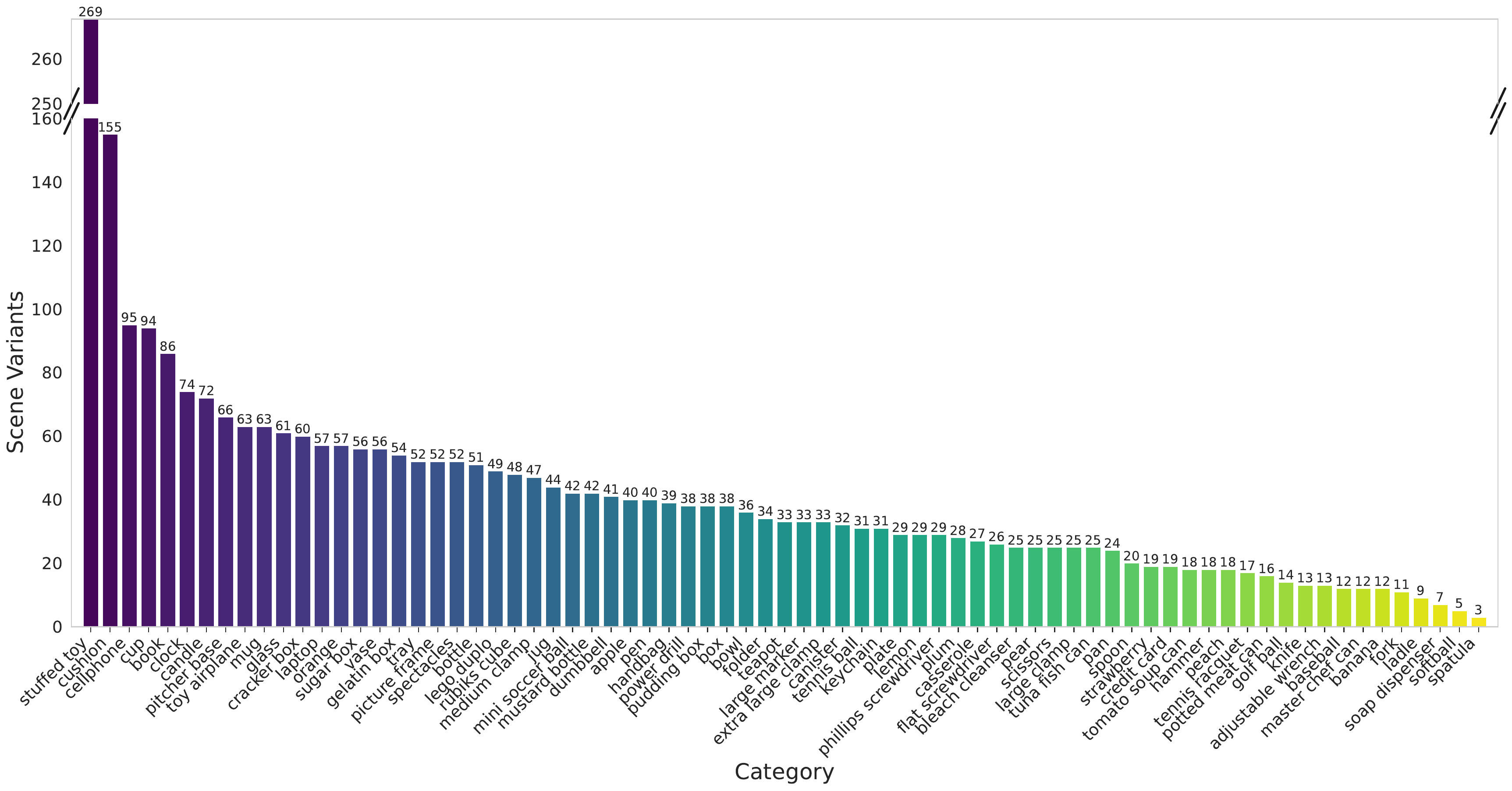}
    \caption{The complete statistics of object category appearance frequencies across the 363 scene variants from the \textit{SD-OVON-3k} dataset.}
    \label{fig:category_scene_stat}
\end{figure}

\begin{figure}[ht]
    \centering
    \includegraphics[width=1.0\linewidth]{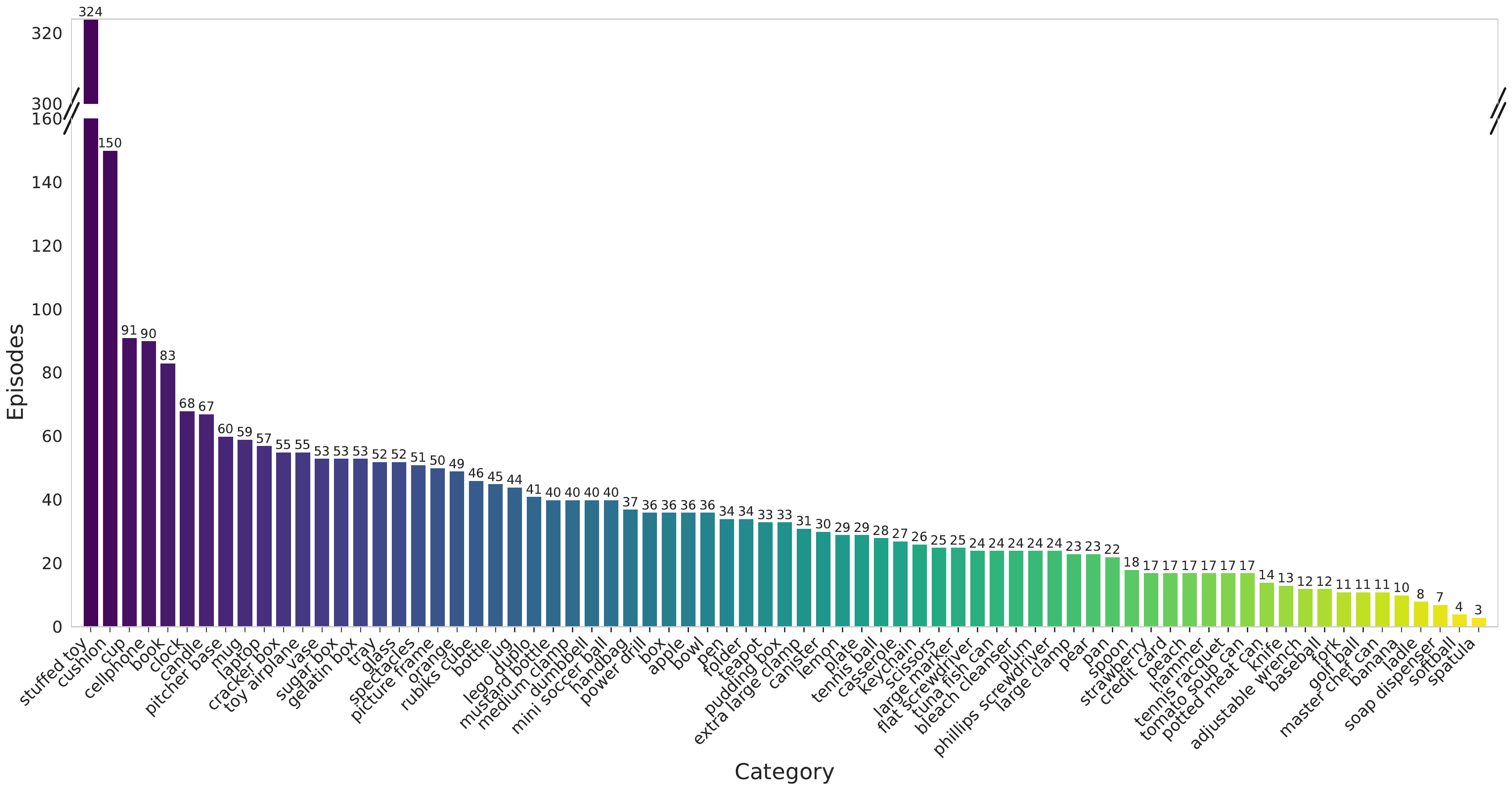}
    \caption{The complete statistics of the navigation goal object category appearance frequencies across the 2897 ObjectNav task episodes from the \textit{SD-OVON-3k} dataset.}
    \label{fig:episode_goal_stat}
\end{figure}

\begin{figure}[ht]
    \centering
    \includegraphics[width=1.0\linewidth]{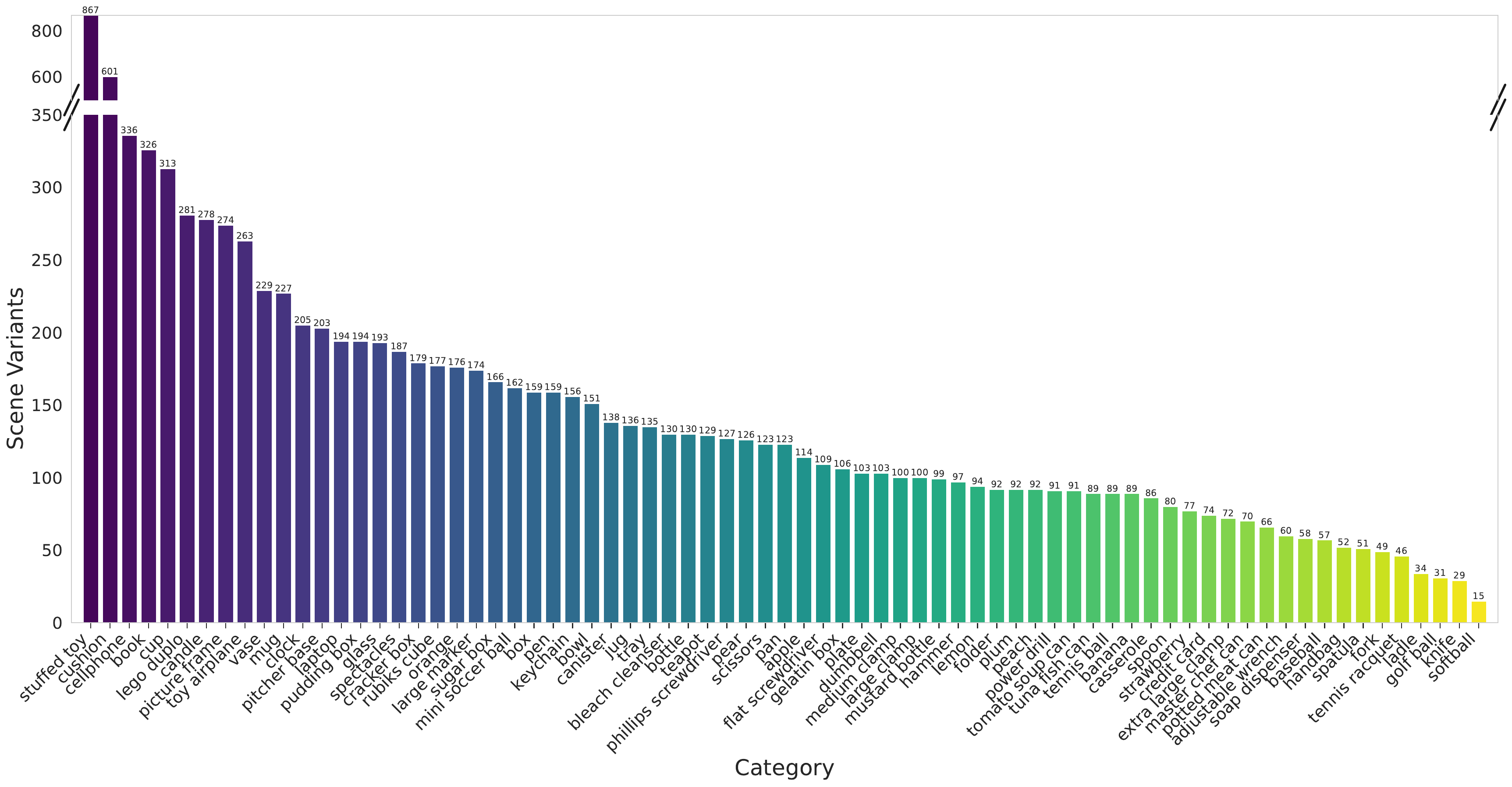}
    \caption{The complete statistics of object category appearance frequencies across the 1127 scene variants from the \textit{SD-OVON-10k} dataset.}
    \label{fig:category_scene_stat_10k}
\end{figure}

\begin{figure}[ht]
    \centering
    \includegraphics[width=1.0\linewidth]{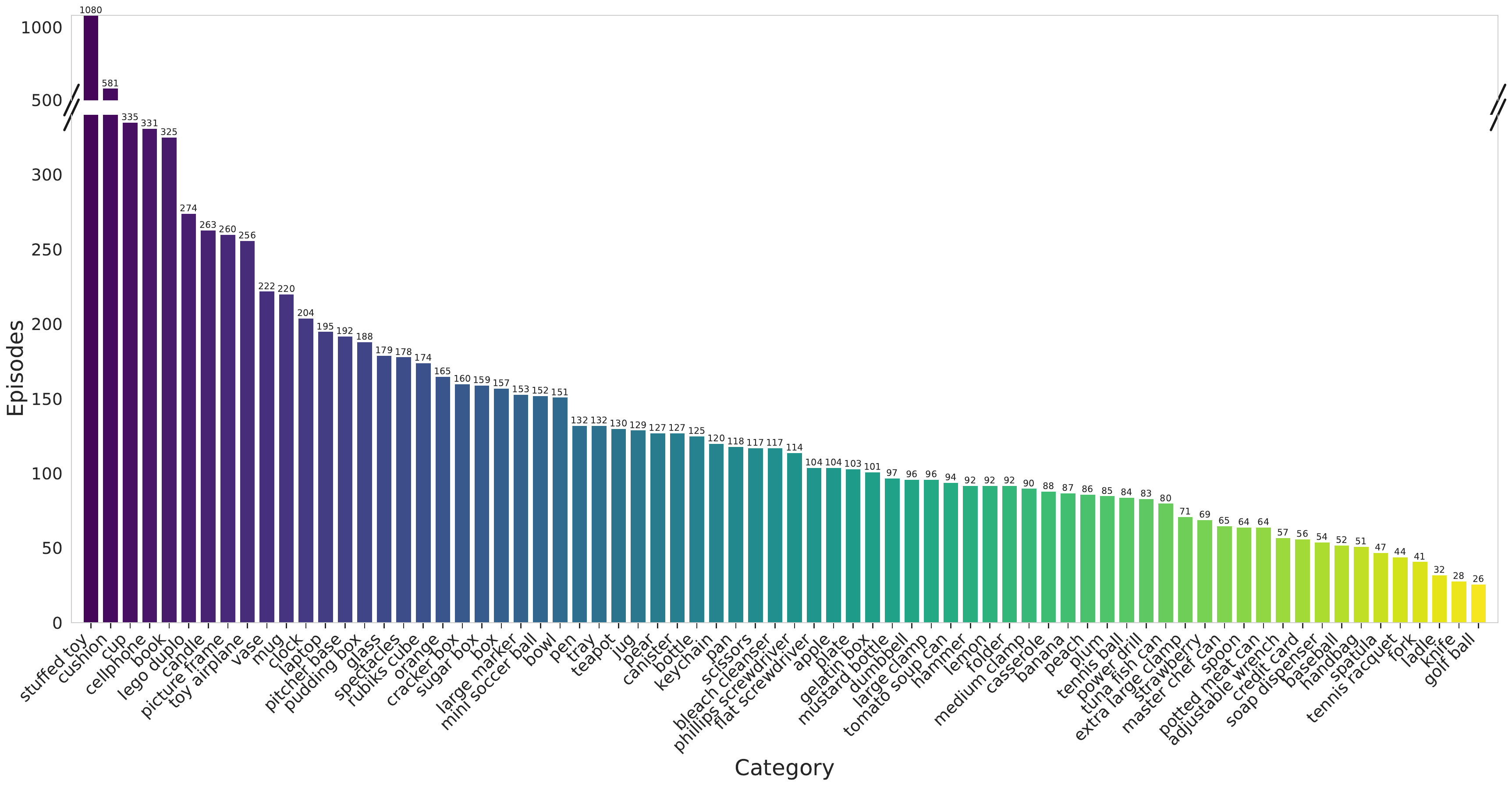}
    \caption{The complete statistics of the navigation goal object category appearance frequencies across the 10629 ObjectNav task episodes from the \textit{SD-OVON-10k} dataset.}
    \label{fig:episode_goal_stat_10k}
\end{figure}

Two accompanied ObjectNav task episodes datasets, SD-OVON-3k and SD-OVON-10k as listed on table \ref{tab:datasets}, are generated  with the SD-OVON pipeline. The complete statistics of object category appearance frequencies across the 363 scene variants and the 2897 ObjectNav task episodes from the SD-OVON-3k dataset are shown in figure \ref{fig:category_scene_stat} and figure \ref{fig:episode_goal_stat}. And the complete statistics of object category appearance frequencies across the 1127 scene variants and the 10629 ObjectNav task episodes from the SD-OVON-10k dataset are shown in figure \ref{fig:category_scene_stat_10k} and figure \ref{fig:episode_goal_stat_10k}.

\section{Uniform Sampling in Convex Polygons}\label{apx:uniform_sampling_in_convex_polygons}

\begin{algorithm}[t]
\caption{Sampling a point in a convex polygon under uniform random distribution}
\label{alg:sampling}
\KwIn{Vertex list $V$ of a convex polygon in counter-clockwise order}
\KwOut{The sampled point $p$}

\SetKwFunction{TrianglePartition}{TrianglePartition}
\SetKwFunction{SelectTriangle}{SelectTriangle}
\SetKwFunction{GenerateSamplePoint}{GenerateSamplePoint}

\BlankLine

\textbf{Step 1:} Construct a triangle partition set $T$ from vertex set $V$ \;
Fix the first vertex $V_1 \in V$ \;
\For{$i = 2$ \KwTo $|V|$}{
    Compute triangle $t_i = (V_1, V_{i-1}, V_i)$ \;
    Compute area $S_{t_i}$ of triangle $t_i$ \;
    Add triangle $t_i$ and its area $S_{t_i}$ into set $T$ \;
}
\textbf{End For}

\BlankLine

\textbf{Step 2:} Uniform sampling \;
\For{$j = 1$ \KwTo $N$}{
    Select triangle $t \in T$ using roulette wheel selection with probability $P(t) = \frac{S_t}{S}$ \;
    
    Set $A = V_1$, $B = V_{i_1}$, $C = V_{i_2}$ for triangle $t$ \;
    Generate two random numbers $r_1, r_2 \sim U(0, 1)$ \;
    
    \If{$r_1 + r_2 \leq 1$}{
        $p = A + r_1 \cdot (B - A) + r_2 \cdot (C - A)$\;
    }
    \Else{
        $p = A + (1 - r_1) \cdot (B - A) + (1 - r_2) \cdot (C - A)$\;
    }
    
    Add the sampled point $p$ to set $P$ \;
}
\textbf{End For}

\BlankLine

\end{algorithm}

\subsection{Uniform Distribution}

We define two independent uniformly distributed random number \(R_1 = r_1\) and \(R_2 = r_2\) on the interval \([0, 1]\).
Then we define the random variable \( R \) as follows:

\[
R = \begin{cases} 
(r_1, r_2) & \text{if } r_1 + r_2 \leq 1, \\
(1 - r_1, 1 - r_2) & \text{if } r_1 + r_2 > 1. 
\end{cases}
\]

First, consider the region where \( r_1 + r_2 \leq 1 \),
\[
\text{Area}(r_1 + r_2 \leq 1) = \frac{1}{2} \times 1 \times 1 = \frac{1}{2}.
\]

Next, consider the region where \( r_1 + r_2 > 1 \),
\[
\text{Area}(r_1 + r_2 > 1) = 1 - \text{Area}(r_1 + r_2 \leq 1) = 1 - \frac{1}{2} = \frac{1}{2}.
\]

For \( (r_1, r_2) \) when \( r_1 + r_2 \leq 1 \): 
\( R \) in this region takes the value \( (r_1, r_2) \). Since \( R_1 \) and \( R_2 \) are uniformly distributed, \( R = ( r_1, r_2) \) is uniformly distributed in area \( r_1 + r_2 \leq 1 \) and its joint probability density function is constant and integrates to \( \frac{1}{2} \).
Thus, the probability density function for this region is:
\[
f_{R}(u, v) = \begin{cases}
1 & \text{if } u + v \leq 1 \text{ and } 0 \leq u, v \leq 1, \\
0 & \text{otherwise}.
\end{cases}
\]

For \( (1 - r_1, 1 - r_2) \) when \( r_1 + r_2 > 1 \): \( R \) in this region takes the value \( (1 - r_1, 1 - r_2) \). First, consider the random variable \( 1 - R_1 \). Since \( 1 - R_1 \) is defined by a linear transformation of \( R_1 \), \( 1 - R_1 \) is uniformly distributed over the interval \( (0, 1) \). Similarly, \( 1 - R_2 \) is also uniformly distributed over the interval \( (0, 1) \).

Therefore, \( R = ( 1 - r_1, 1 - r_2)\) is uniformly distributed in area \( r_1 + r_2 > 1 \) and its joint probability density function is constant and integrates to \( \frac{1}{2} \).

Thus, the probability density function for this region is:
\[
f_{R}(u, v) = \begin{cases}
1 & \text{if } u + v \geq 1 \text{ and } 0 \leq u, v \leq 1, \\
0 & \text{otherwise}.
\end{cases}
\]

Combining these two parts, the joint probability density function of the random variable \( R \) is:
\[
f_R(u, v) = \begin{cases}
1 & \text{if } 0 \leq u, v \leq 1, \\
0 & \text{otherwise}.
\end{cases}
\]

This indicates that \( R \) is uniformly distributed over the unit square \( (0,1) \times (0,1) \).

\subsection{Parameterizing the Triangle}

Consider a triangle defined by its vertices \(A\), \(B\), \(C\). Let \(\mathbf{v} = \overrightarrow{AB}\) and \(\mathbf{w} = \overrightarrow{AC}\). Any point \(\mathbf{P}\) within the triangle can be expressed as:
\[
\mathbf{P}(s, t) = \mathbf{A} + s \mathbf{v} + t \mathbf{w}.
\]
where \( (s, t) \) is generated by values of two independent uniformly distributed random variables \(R_1 = r_1, R_2 = r_2\) on the interval \([0, 1]\) as follows:
\[
(s, t) = \begin{cases} 
(r_1, r_2) & \text{if } r_1 + r_2 \leq 1, \\
(1 - r_1, 1 - r_2) & \text{if } r_1 + r_2 > 1. 
\end{cases}
\]
The above proof indicates that \( (s, t) \) is also uniformly distributed over the region \( 0 \leq s, t \leq 1\) and \( 0 \leq s + t \leq 1\).

\subsection{Uniform Sampling}

Since the sample point \(\mathbf{P}(s, t)\) is defined by a linear transformation from the parameter space to Euclidean space, this linear transformation preserves the uniform distribution property of \(\mathbf{P}(s, t)\) within the triangle from the parameter space to Euclidean space, ensuring that the sampled point is uniformly sampled within the triangle in Euclidean space. Furthermore, since the probability of selecting each triangle $t$ is $P(t) = \frac{S_t}{S}$ and $\sum S_t = S$ and the sampling within each selected triangle is uniform, the distribution of sample points over the entire polygon is also uniform. Algorithm \ref{alg:sampling} depicts the complete procedure of uniformly sampling a point in a convex polygon.

\section{LLM Prompts Design}

The following subsections list the LLM prompts designed for modules that involve LLMs.

\subsection{Prompt for Identifying Receptacles}\label{apx:llm_prompt_identifying_receptacles}

\begin{leftbubblessingle}{bubblegray}{System}{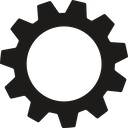}
    The user will give you a list of objects \{"objects": [object\_1, object\_2, ...]\}, please find and return pieces of furnitures with surface to put daily objects on top of, such as "sofa", "bed", and "table", which we refer to as "receptacles". Return in JSON format \{"receptacles": [receptacle\_1, receptacle\_2, ...]\}. Do NOT reply any other information.
\end{leftbubblessingle}

\begin{rightbubblessingle}{bubblegreen}{Input}{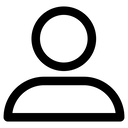}
    \{"objects": "curtain", "bed", "stool", "pillow", "desk", "shelf", "book shelf", "dresser", "chair", "pen", "laptop", "lamp"\}
\end{rightbubblessingle}

\begin{leftbubblessingle}{blue!10}{LLM}{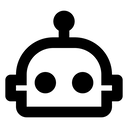}
    \{"receptacles": ["bed", "desk", "shelf", "book shelf", "dresser", "chair"]\}
\end{leftbubblessingle}

\subsection{Prompt for Region Proposals}\label{apx:llm_prompt_region_proposals}

\begin{leftbubblessingle}{bubblegray}{System}{figures/system_icon.png}
    The user will give you a list of objects \{"objects": [object\_1, object\_2, ...]\} in a region inside a house or a public indoor space, please suggest what this region may be. Propose at least five candidates, only return JSON format \{"regions": [region\_1, region\_2, ...]\}. Do NOT reply other information.
\end{leftbubblessingle}

\begin{rightbubblessingle}{bubblegreen}{Input}{figures/human_icon.png}
    \{"objects": ["headboard", "bed", "stool", "carpet", "pillow", "curtain"]\}
\end{rightbubblessingle}

\begin{leftbubblessingle}{blue!10}{LLM}{figures/robot_icon.png}
    \{"regions": ["Bedroom", "Living Room", "Dining Room", "Home Office", "Guest Bedroom"]\}
\end{leftbubblessingle}

\subsection{Prompt for Object-receptacle Relevance Scoring}\label{apx:llm_prompt_receptacle_relevance}

\begin{leftbubblessingle}{bubblegray}{System}{figures/system_icon.png}
    The user will give you a list of pieces of furniture and an object. Give a relevance score from 0 (low) to 10 (high) for each piece of furniture indicating how likely the given object may be placed on it in daily situations, and tell your reason in details. Return in JSON format only: \{"reason": reason, "relevances": \{furniture\_name: score, ...\}\}. Do NOT reply other information.
\end{leftbubblessingle}

\begin{rightbubblessingle}{bubblegreen}{Input}{figures/human_icon.png}
    \{"furniture": ["table", "desk", "dresser", "bookshelf", "shelf", "bed", "sofa", "couch"], "object": "dumbbell"\}
\end{rightbubblessingle}

\begin{leftbubblessingle}{blue!10}{LLM}{figures/robot_icon.png}
    \{"reason": The dumbbell is likely to be placed on a surface where it can be easily accessed and used. "relevances": \{"table": 8, "desk": 9, "dresser": 1, "bookshelf": 0, "shelf": 4, "bed": 2, "sofa": 5, "couch": 6\}\}
\end{leftbubblessingle}

\subsection{Prompt for Object-region Relevance Scoring}\label{apx:llm_prompt_region_relevance}

\begin{leftbubblessingle}{bubblegray}{System}{figures/system_icon.png}
    The user will give you a list of regions and an object. Give a score from 0 (low) to 10 (high) for each region indicating the relevance between this region and the given object, after telling your reason in details. Return in JSON format only: \{"reason": reason, "relevances": \{region\_name: score, ...\}\}. Do NOT reply other information.
\end{leftbubblessingle}

\begin{rightbubblessingle}{bubblegreen}{Input}{figures/human_icon.png}
    \{"regions": ["living room", "living room and kitchen combination", "bedroom", "bathroom", "kitchen"], "object": "dumbbell"\}
\end{rightbubblessingle}

\begin{leftbubblessingle}{blue!10}{LLM}{figures/robot_icon.png}
    \{"reason": "The score is based on the likelihood of finding the object in a specific region, considering its typical uses and associations.", "relevances": \{"bedroom": 8, "living room": 6, "living room and kitchen combination": 7, "bathroom": 1, "kitchen": 2\}\}
\end{leftbubblessingle}

\subsection{Restricted LLM Output Formatting}

We restrict the output from LLMs formatted in JSON. Structural LLM outputs facilitate accurate data parsing. Each output from LLMs will be checked for format, and re-generated if any it is not in standard JSON format. And an example input to and the corresponding output from the LLM \cite{grattafiori2024llama} we adopt in our implementation is provided for each of the LLM prompts.

\section{Implementations of Baselines}\label{apx:baselines}

The implementations of the Random Receptacle Navigation A* and the Semantic Navigation A* are discussed in details in the following subsections.

\subsection{Random Receptacle Navigation A*}\label{apx:random_astar_details}

The Random Receptacle Navigation A*, presented in algorithm \ref{alg:random_astar}, preserves prior knowledge of the environment semantics during its initial visit to a scene. It explores the scene with frontier-based methods~\cite{sun2020frontier, cao2022autonomous} ("\texttt{Explore}") when it visits a scene for the first time. Instance extraction and fusion (as introduced in section \ref{sec:Instance_Extraction_and_Fusion}) and receptacle planes identifying (as introduced in section \ref{sec:Receptable_Planes_Identifying}) follow to extract each receptacle plane $S$ with receptacle category semantics $l_{\text{rec}}$ ("\texttt{ExtractPlanes}") from the observations accumulated during exploration. Dense navigation points can be sampled ("\texttt{SampleNavpoints}") surrounding extract receptacle planes in advance, which are navigable positions for the agent to observe the objects placed on the receptacle.
At each following episode, the order of receptacles to visit are randomly shuffled ("\texttt{Shuffle}"). The agent uses the A* planner for point navigation ("\texttt{Navigate}") with error tolerance $\epsilon_{\text{nav}}$ towards intermediate navigation points and the goal object if it is found. GroundingDINO~\cite{liu2023grounding} is adopted for open-vocabulary object detection ("\texttt{Detect}") at each navigation point. And equations \ref{eq:reprojection} and \ref{eq:instance_points_decomposition} are used to reconstruct ("\texttt{Reconstruct}") the 3D position of the goal object for final point goal navigation towards it.

\begin{algorithm}[t]
\caption{Random Receptacle Navigation A*}
\label{alg:random_astar}
\KwIn{scene $s$, goal object $g$, initial agent position $p_{0}$, memory $M$, navigation error tolerance $\epsilon_{\text{nav}}$, minimum next navigation point distance $d_{\text{next}}$}
\KwOut{navigation trajectory $\tau$, memory $M$}
\SetKw{KwReturn}{return}
\SetKwFunction{KwFuncIndices}{Indices}
\SetKwFunction{KwFuncExplore}{Explore}
\SetKwFunction{KwFuncExtractPlanes}{ExtractPlanes}
\SetKwFunction{KwFuncSampleNavpoints}{SampleNavpoints}
\SetKwFunction{KwFuncDetect}{Detect}
\SetKwFunction{KwFuncReconstruct}{Reconstruct}
\SetKwFunction{KwFuncShuffle}{Shuffle}
\SetKwFunction{KwFuncNavigate}{Navigate}
\BlankLine
\If{$s \notin \KwFuncIndices(M)$}{
    Explore the environment $\tau_{0} \sim \KwFuncExplore(s)$\;
    Accumulate observations $L_{\text{obs}} = \{ \langle \mathcal{I}, \mathcal{D}, \mathbf{T}_{\text{cam}} \rangle \} \sim O(\tau_{0}|s) $\;
    Extract receptacle planes with semantics $L_{\text{RP}} = \{ \langle S, l_{\text{rec}}^{S} \rangle \} \gets \KwFuncExtractPlanes(L_{\text{obs}})$\;
    Sample navigation points $P_{\text{nav}}^{S} = \{ p_{\text{nav}}^{S} \} \sim \KwFuncSampleNavpoints(S), \forall S \in L_{\text{RP}} $\;
    Update memory $M(s) \gets \{ \langle S, l_{\text{rec}}^{S}, P_{\text{nav}}^{S} \rangle \}$\;
}
Initialize $\tau \gets \{ p_{0} \}$\;
Randomly shuffle visit order of receptacles $V \sim \KwFuncShuffle(M(s))$\;
\For{$\langle S, l_{\text{rec}}, P_{\text{nav}} \rangle \in V$}{
    \For{$p_{\text{nav}} \in P_{\text{nav}}$}{
        Plan towards next navigation point $\langle p, \tilde{\tau} \rangle \gets \KwFuncNavigate(p_{\text{nav}} | \epsilon_{\text{nav}})$ with A*\;
        Update $\tau \gets \tau + \tilde{\tau}$\;
        Observe $o \sim O(p|s)$\;
        \If{$g \in \KwFuncDetect(o)$}{
            Reconstruct 3D goal object position $p_{g} \gets \KwFuncReconstruct(g|o)$\;
            Plan towards goal $\langle p, \tilde{\tau} \rangle  \gets \KwFuncNavigate(p_{g} | \epsilon_{\text{nav}}) $ with A*\;
            \KwReturn $\tau$, $M$\;
        }
        Skip nearby navigation points $P_{\text{nav}} \gets \{ p' \in P_{\text{nav}} | \|p' -p\| \geq d_{\text{next}} \}$\;
    }
}
\KwReturn $\tau$, $M$\;
\end{algorithm}

\subsection{Semantic Navigation A*}\label{apx:semantic_astar_details}

The Semantic Navigation A* adopts similar procedures to the Random A* algorithm. However, it extracts not only the receptacle semantics but also the region semantics to build region-receptacle semantic awareness of the environment during the initial exploration stage for later on prioritization of the receptacles to visit in accord to region-receptacle semantics.

Semantic A*, presented in algorithm \ref{alg:semantic_astar}, explores the scene with frontier-based methods~\cite{sun2020frontier, cao2022autonomous} ("\texttt{Explore}") when it visits a scene for the first time. Instance extraction, fusion and region-receptacle semantics inference (as introduced from section \ref{sec:Instance_Extraction_and_Fusion} to section \ref{sec:Semantic_Object_Placement}) follow to extract each receptacle plane $S$ with receptacle category semantics $l_{\text{rec}}$ and region semantics $l_{\text{rgn}}$ ("\texttt{ExtractPlanes}") from the observations accumulated during exploration. Dense navigation points can be sampled ("\texttt{SampleNavpoints}") surrounding extract receptacle planes in advance, which are navigable positions for the agent to observe the objects placed on the receptacle.
At each following episode, it prioritizes the receptacles to visit by decreasing order of relevance scores of the target object to region-receptacle pairs following equation \ref{eq:rro_joint_relevance}. The agent then uses the A* planner for point navigation ("\texttt{Navigate}") with error tolerance $\epsilon_{\text{nav}}$ towards intermediate navigation points and the goal object if it is found. GroundingDINO~\cite{liu2023grounding} is used for open-vocabulary object detection ("\texttt{Detect}") at each navigation point. And equations \ref{eq:reprojection} and \ref{eq:instance_points_decomposition} are used to reconstruct ("\texttt{Reconstruct}") the 3D position of the goal object for final point goal navigation towards it.

\begin{algorithm}[t]
\caption{Semantic Navigation A*}
\label{alg:semantic_astar}
\KwIn{scene $s$, goal object $g$, initial agent position $p_{0}$, memory $M$, navigation error tolerance $\epsilon_{\text{nav}}$, minimum next navigation point distance $d_{\text{next}}$}
\KwOut{navigation trajectory $\tau$, memory $M$}
\SetKw{KwReturn}{return}
\SetKwFunction{KwFuncIndices}{Indices}
\SetKwFunction{KwFuncExplore}{Explore}
\SetKwFunction{KwFuncExtractPlanes}{ExtractPlanes}
\SetKwFunction{KwFuncSampleNavpoints}{SampleNavpoints}
\SetKwFunction{KwFuncDetect}{Detect}
\SetKwFunction{KwFuncReconstruct}{Reconstruct}
\SetKwFunction{KwFuncShuffle}{Shuffle}
\SetKwFunction{KwFuncNavigate}{Navigate}
\BlankLine
\If{$s \notin \KwFuncIndices(M)$}{
    Explore the environment $\tau_{0} \sim \KwFuncExplore(s)$\;
    Accumulate observations $L_{\text{obs}} = \{ \langle \mathcal{I}, \mathcal{D}, \mathbf{T}_{\text{cam}} \rangle \} \sim O(\tau_{0}|s) $\;
    Extract receptacle planes with semantics $L_{\text{RP}} = \{ \langle S, l_{\text{rgn}}^{S}, l_{\text{rec}}^{S} \rangle \} \gets \KwFuncExtractPlanes(L_{\text{obs}})$\;
    Sample navigation points $P_{\text{nav}}^{S} = \{ p_{\text{nav}}^{S} \} \sim \KwFuncSampleNavpoints(S), \forall S \in L_{\text{RP}} $\;
    Update memory $M(s) \gets \{ \langle S, l_{\text{rgn}}^{S}, l_{\text{rec}}^{S}, P_{\text{nav}}^{S} \rangle \}$\;
}
Initialize $\tau \gets \{ p_{0} \}$, $V \gets M(s)$\;
\For{$v = \langle S, l_{\text{rgn}}, l_{\text{rec}}, P_{\text{nav}} \rangle \in V, \langle l_{\text{rgn}}, l_{\text{rec}} \rangle = \arg\max \mathbb{P}(g|l_{\text{rgn}}, l_{\text{rec}}) $}{
    \For{$p_{\text{nav}} \in P_{\text{nav}}$}{
        Plan towards next navigation point $\langle p, \tilde{\tau} \rangle \gets \KwFuncNavigate(p_{\text{nav}} | \epsilon_{\text{nav}})$ with A*\;
        Update $\tau \gets \tau + \tilde{\tau}$\;
        Observe $o \sim O(p|s)$\;
        \If{$g \in \KwFuncDetect(o)$}{
            Reconstruct 3D goal object position $p_{g} \gets \KwFuncReconstruct(g|o)$\;
            Plan towards goal $\langle p, \tilde{\tau} \rangle  \gets \KwFuncNavigate(p_{g} | \epsilon_{\text{nav}}) $ with A*\;
            \KwReturn $\tau$, $M$\;
        }
        Skip nearby navigation points $P_{\text{nav}} \gets \{ p' \in P_{\text{nav}} | \|p' -p\| \geq d_{\text{next}} \}$\;
    }
    Update $V \gets V - \{v\}$\;
}
\KwReturn $\tau$, $M$\;
\end{algorithm}

\section{List of Hyperparameters}\label{apx:list_of_hyperparameters}

Table \ref{tab:list_of_hyperparameters} lists the hyperparameters and their default values used in our implementation.

\begin{table}[t]
\centering
\caption{List of hyperparameters.}
\label{tab:list_of_hyperparameters}
\begin{tabularx}{\textwidth}{cXc}
    \toprule
    Hyperparameter & Description & Default Value \\
    \midrule
    $h_{\text{map}}$ & Height above the floor to slice for top-down map & 0.3 \\
    $\epsilon_{p_{\text{obs}}}$ & Strong observation threshold& 0.5 \\
    $k_{sem}$ & Semantic weight in instance fusion & 0.4 \\
    $k_{geo}$ & Geometric weight in instance fusion & 1.6 \\
    $\varphi_{\min}$ & Minimum successful match similarity in instance fusion & 0.8 \\
    $\text{IoU}_{\min}$ & Minimum ratio of intersection over union (IoU) between the 3D bounding boxes of two instances to consider them significantly overlapped with each other & 0.9 \\
    $h_{\text{spawn}}$ & Height above the receptacle surface to spawn object instances before physics simulation in object placement & 0.3 \\
    $\eta_{\text{gt}}$ & Minimum ratio of pixel sizes of the object over the camera to consider an object being observed in the ground-truth semantics detector (GT) & 0.0001 \\
    $\epsilon_{\text{nav}}$ & Navigation error tolerance for Random A* and Semantic A* & 0.5 \\
    $d_{\text{next}}$ & Minimum distance between the agent and the next navigation point to search in Random A* and Semantic A* & 1.0 \\
    \bottomrule
\end{tabularx}
\end{table}

\section{Compute Resources}

All our experiments, including the pre-generation of the datasets and the evaluation of baselines, are conducted on a workstation with 2 NVIDIA GeForce RTX 4090 GPUs, a 32-core 13th Gen Intel(R) Core(TM) i9-13900K CPU with 128GB flash memory. We also tested our pipeline and the baselines on a workstation with only a single NVIDIA GeForce RTX 4090 GPU.

\end{document}